\documentclass{ifm-tr}

\usepackage{amsmath,amsfonts,bm}

\def\eqref#1{equation~\ref{#1}}

\def\1{\bm{1}}

\DeclareMathAlphabet{\mathsfit}{\encodingdefault}{\sfdefault}{m}{sl}
\SetMathAlphabet{\mathsfit}{bold}{\encodingdefault}{\sfdefault}{bx}{n}

\usepackage{amsmath,amssymb,amsfonts,amsthm,mathrsfs}
\usepackage{xspace,textcomp,soul,booktabs,makecell,multirow}
\usepackage{algorithm,algorithmicx,algpseudocode,listings}
\usepackage{enumitem,float,longtable}
\usepackage{tikz}
\usetikzlibrary{calc}

\graphicspath{{figures/}{assets/}}
\newcommand{\modelname}{PAN\xspace}
\title{PAN: A World Model for General, Actionable, and Long-Horizon World Simulation}

\author[]{\textbf{PAN Team, Institute of Foundation Models}}\authorsep{}
\affiliation[]{Mohamed bin Zayed University of Artificial Intelligence}
\reportnumber{2025-11-14}

\abstract{
A world model is a cognitive simulator of the real-world environment allowing biological agents to reason about how the world evolves, whether spontaneously or in response to their actions, and accordingly to plan and strategize.
In building Artificial Intelligence (AI) systems, world models represent the next frontier beyond large language models (LLMs) to enable physical and embodied intelligence in AI agents, allowing them to perform decision-making through simulative reasoning and reinforcement-learning through simulative trials.
Recent advancements in world modeling have yielded impressive progress in video generation, 3-D scene evolution, robotic dynamics, and game simulation,
but limitations persist in general, open-domain, action-driven prediction, long-horizon consistency, and abstract reasoning and planning.
Moreover, fundamental architectural questions, whether it be state representation, information flow, or training objectives, remain unresolved.
In this paper, we introduce PAN, a world model built on the Generative Latent Prediction (GLP) architecture.
GLP combines stateful latent representations of world states; an encoder--decoder closed-loop information flow; an LLM/diffusion-based mixed reasoning backbone; and a non-degenerate generative reconstruction objective whose fidelity is ``dampable'' to balance fine-grained detail against semantic saliency.
Compared to several existing systems, PAN demonstrates advantages beyond standard video generation in action-conditioned world simulation, long-horizon forecasting, and simulative reasoning and planning, capabilities we argue should serve as the primary criteria for evaluating world models.
}

\begin{document}
\maketitle

\addtocontents{toc}{\protect\setcounter{tocdepth}{-10}}

\section{Introduction}\label{sec1}

The study of intelligent decision-making, from formulation of model-based reinforcement learning~\citep{sutton2017reinforcement} to framework of universal intelligence~\citep{legg2007universal}, converges on a common insight: an agent that can predict how its environment responds to actions holds a decisive advantage in planning and control.
Formally, this amounts to learning a \emph{next-state-prediction} function, namely a conditional distribution $p(s' \mid s, a)$ that, given the current world state $s$ and a candidate action $a$, forecasts the resulting state $s'$.
A \emph{world model} is precisely such a learned simulator.
By querying it with hypothetical actions, an agent can perform ``thought experiments,'' evaluating the consequences of alternative plans before committing to any of them, a capacity that the psychology literature terms \emph{hypothetical thinking}~\citep{Ball_2020} and that underpins general intelligence~\citep{xing2025critiques}.

Classical world models in reinforcement learning, whether it be early learned dynamics models~\citep{ha2018world}, the Dreamer series~\citep{hafner2019dream,hafner2023mastering}, or discrete-token predictors~\citep{micheli2022transformers}, demonstrated that learned next-state prediction can dramatically improve sample efficiency and planning.
However, these models are trained and evaluated within single task-specific environments with narrow, predefined action spaces, limiting both their domain coverage and their generality as reasoning tools.
A parallel line of work has explored using large language models (LLMs) as flexible world models for reasoning and planning~\citep{xiang2023language,hao2023reasoning,deng2025general}, but LLMs are not natively configured to learn from non-linguistic data, and their world knowledge remains limited to what can be expressed and acquired through language.
Meanwhile, the emergence of large-scale video generation models~\citep{videoworldsimulators2024,veo2024,yang2024cogvideox,kong2024hunyuanvideo,wan2025wan} sparked widespread interest in whether such models could serve as general world simulators that go beyond simple game or text environments into broader physical realms.
The visual fidelity of such models is remarkable, but visual realism alone does not constitute world modeling.
These systems are typically optimized for open-loop media generation, producing complete videos from fixed prompts without maintaining persistent internal states, exposing controllable action interfaces, or representing object-level causal dynamics, all of which are essential for purposeful reasoning, counterfactual imagination, and decision evaluation.

On the other hand, existing world models make different architectural choices, each with characteristic strengths and trade-offs.
Domain-specific world models for gaming~\citep{micheli2022transformers,alonso2024diffusion,bruce2024genie}, autonomous driving~\citep{wang2023driving,hu2023gaia}, and robotics~\citep{yang2024learning,cosmos} achieve strong within-domain performance but rely on restricted action spaces and task-specific design that limit generality.
Latent predictive models such as JEPA~\citep{lecun2022path,assran2025vjepa2selfsupervisedvideo} avoid pixel-level reconstruction by predicting future states in an embedding space, but their purely latent-space objectives can admit degenerate solutions~\citep{xing2025critiques} and may discard information needed for faithful simulation: visual details that are weakly expressed in the target representation are compressed away even when they are dynamically and logically relevant.
Moreover, restricting the backbone to continuous embeddings forgoes the benefits of symbolic reasoning, limiting long-horizon consistency and compositional or counterfactual generalization.
3D world models~\citep{worldlabs2024generating} capture spatial and geometric structure but typically lack temporal dynamics, multi-agent interaction, or meaningful actions beyond navigation.
The challenge of combining open-domain visual coverage with action-conditioned temporal consistency, while maintaining observation-grounded dynamic rollout, remains open.

In this work, we introduce \textbf{PAN}, a general world model built on the \textbf{Generative Latent Prediction (GLP)} architecture~\citep{xing2025critiques}.
GLP addresses the fundamental tension in world model design between \emph{latent abstraction}, which enables scalable, long-horizon state evolution, and \emph{observational grounding}, which ensures that predicted transitions correspond to realizable sensory changes.
It does so through three coupled components: an \emph{encoder}~$h$ that maps observations into a structured latent state, a \emph{predictive backbone}~$f$ that evolves latent states forward under actions, and a \emph{decoder}~$g$ that reconstructs observable outcomes from predicted next states.
The decoder is not merely a visualization module, but the training signal.
By supervising predictions through a generative reconstruction objective in observation space, GLP avoids the representation collapse and information loss that afflict purely latent-space objectives~\citep{xing2025critiques}. Specifically, the encoder is continuously pressured to retain all dynamically relevant information, because any discarded detail will surface as reconstruction error.
This reconstruction-based design also distinguishes GLP from embedding-prediction frameworks such as JEPA, whose latent-space losses are, at best, loose surrogates for observation-level consistency and whose regularizers introduce optimization complexity without guaranteeing grounding.
By separating abstract causal dynamics (modeled in the backbone) from fine-grained perceptual realization (handled by the decoder), GLP offers a principled, scalable foundation for general, interactive, and causally grounded world models.

\modelname makes deliberate design choices in implementing the GLP architecture, each addressing a specific challenge that the architecture faces in practice.
To overcome the information sparsity of raw perceptual data, PAN adopts a large language model (LLM) as the world model reasoning backbone, inheriting broad multimodal knowledge and language-conditioned sequence modeling from pretraining; this provides the symbolic reasoning capacity that continuous-only backbones lack, supporting compositional and logical (such as counterfactual) generalization and long-horizon consistency.
A common objection to generative world models is that reconstructing all details of the next observation is intractable~\citep{lecun2022path}. PAN addresses this through the \textbf{Causal Shift-Window Denoising Process Model (Causal Swin-DPM)}, which augments the denoising process of~\citet{feng2024matrix} with a chunk-wise causal attention mask. By conditioning each chunk on partially noised, rather than fully denoised, representations of preceding chunks, the mechanism dampens incidental pixel-level details and focuses supervision on semantically persistent structure, effectively modulating the trade-off between fidelity and saliency; this makes generative reconstruction practical without requiring the backbone to determine every fine-grained detail.
Finally, to provide the action-conditioned diversity necessary for general-purpose simulation, PAN is trained on 8 million video--text pairs from publicly accessible sources, with dense captions emphasizing temporal dynamics rather than static scene attributes. Together, these choices yield an open-domain, language-conditioned world model with coherent multi-step dynamics; qualitative examples of multi-step interactive rollouts over 10+ action steps are shown in Extended Data Fig.~1.

We evaluate \modelname on the World Reasoning Arena (WR-Arena) benchmark~\citep{gao2026wrarena}, which measures world model competence along three complementary dimensions: action simulation fidelity, long-horizon forecasting, and simulative reasoning and planning (see \S\ref{sec:results} and Supplementary Information \S4 for full benchmark specifications).
\modelname achieves the highest overall performance among open-source world models and is competitive with leading commercial systems on action-conditioned simulation and long-horizon forecasting.
When integrated with a VLM-based planning agent, \modelname yields consistent improvements in task success over both the agent-only baseline and agent--baseline-model combinations.
To isolate the contributions of PAN's key components, we conduct a progressive ablation decomposing the system into its base video decoder, the Causal Swin-DPM training framework, and the VLM-based latent dynamics backbone (Supplementary Information \S6).
We discuss remaining limitations, including the constraints of language-level action interfaces and long-horizon drift, in \S\ref{sec:discussion}.
To support reproducibility and further research, we will release \modelname model weights, code, and the WR-Arena evaluation suite.

\section{PAN architecture}\label{sec:arch}

\subsection{Generative latent prediction}

\modelname employs the GLP architecture for world models~\citep{xing2025critiques}, which unifies two complementary requirements of a world model:
\emph{latent dynamics} for scalable, long-horizon state evolution under actions; and
\emph{generative supervision} so that latent transitions correspond to evidence-based sensory changes.
Formally, given an observation $o_t$ and action $a_t$, GLP introduces a latent world state $\hat{s}_t$ and predicts the next observation $\hat{o}_{t+1}$ through three components:\footnote{We adopt the Markov model here for clarity, but it is formally possible to represent non-Markovian dynamics by defining an augmented latent state $\tilde{s}_t = [\hat{s}_1, a_1, \hat{s}_2, \dots, a_{t-1},\hat{s}_t]$ which concatenates the states and actions of all previous time steps. }
\begin{align}
\textbf{Encoder } h: \quad & \hat{s}_t \sim p_h(\cdot \mid o_t), \label{eq:glp-enc}\\
\textbf{Predictive Backbone } f: \quad & \hat{s}_{t+1} \sim p_f(\cdot \mid \hat{s}_t, a_t), \label{eq:glp-dyn}\\
\textbf{Decoder } g: \quad & \hat{o}_{t+1} \sim p_g(\cdot \mid \hat{s}_{t+1}). \label{eq:glp-dec}
\end{align}

\begin{figure}
    \centering
    \includegraphics[width=0.85\linewidth]{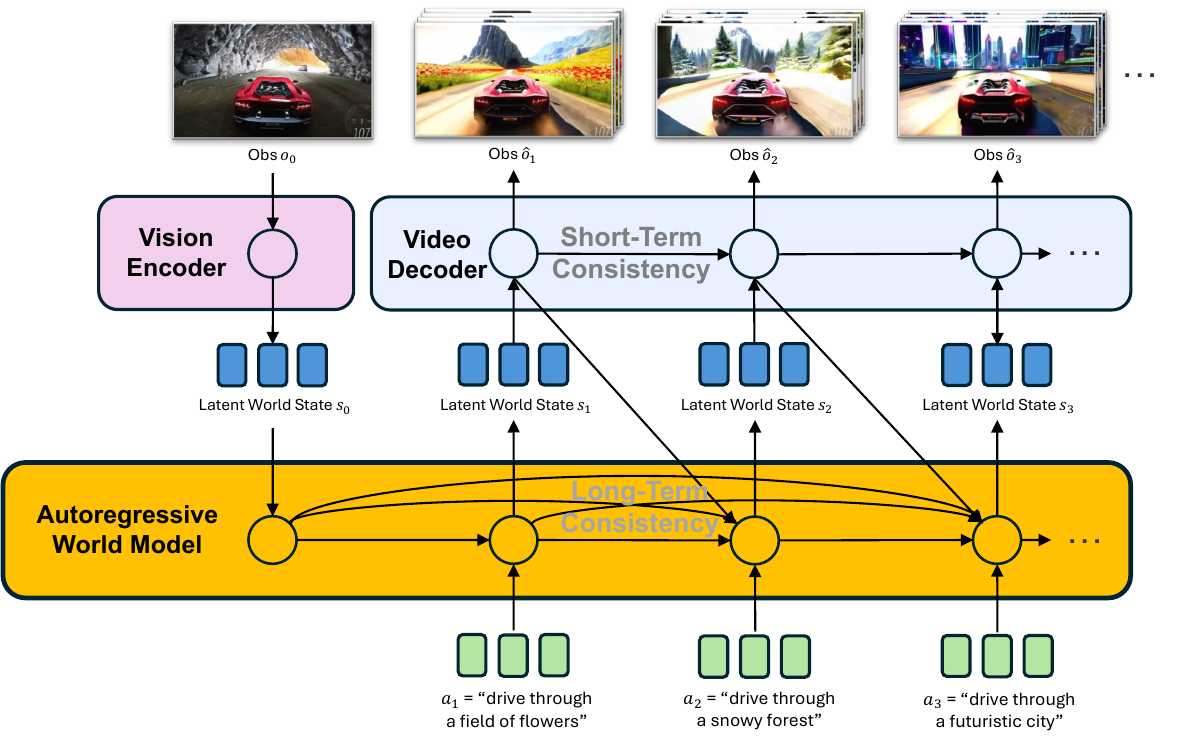}
    \caption{\textbf{PAN model architecture.} The system consists of an autoregressive LLM-based world model backbone for long-horizon world simulation, and a video diffusion decoder for observation prediction. The vision encoder maps observations to latent states; the backbone predicts the next latent state conditioned on state--action history; the decoder renders predictions into temporally coherent video segments via Causal Swin-DPM. Full architectural details are in Supplementary Information \S1.}
    \label{fig:model-arch}
\end{figure}

In this minimal formulation, the decoder depends only on the predicted latent state.
In \modelname, a timestep corresponds to a short observation segment rather than an isolated image frame.
Specifically, each observation $o_t$ denotes a video chunk represented in the latent space of a causal video VAE with spatio-temporal downsampling.
The encoder maps this observation chunk into a latent world state, the predictive module evolves the latent state over time under the history of previous states and actions, and the decoder renders the predicted latent state back into an observable video chunk.
This chunk-level formulation allows temporal structure to be modeled both in the autoregressive latent dynamics across steps and in the video diffusion decoder within and across adjacent chunks.
In \modelname's implementation, the decoder additionally conditions on the most recent observation context through the Causal Swin-DPM sliding window mechanism, as reflected in the operational rollout equations below:
\begin{align}
    \hat{s}_1 &= h(o_1), \label{eq:rollout-encode} \\
    \hat{s}_{t+1} &\sim p_f(\cdot \mid \hat{s}_{\leq t}, a_{\leq t}), \label{eq:rollout-predict}\\
    \hat{o}_{t+1} &\sim p_g(\cdot \mid \hat{s}_{t+1}, a_t, \hat{o}_{t}). \label{eq:rollout-decode}
\end{align}
This additional conditioning provides local visual continuity cues that complement the latent state, but the latent state $\hat{s}_{t+1}$ remains the primary carrier of action-conditioned dynamics information.

\subsection{Generative supervision}

\modelname is trained on tuples $(o_t,a_t,o_{t+1})$ with a reconstruction-based objective:
\begin{align}
\mathcal{L}_{\mathrm{PAN}} &= \mathbb{E}_{(o_t,a_t,o_{t+1})}\Big[\mathrm{disc}(\hat{o}_{t+1},o_{t+1})\Big], \label{eq:loss}
\end{align}
where $\hat{o}_{t+1} = g(f(h(o_t),a_t))$.
This generative supervision keeps latent dynamics grounded: each transition from $\hat{s}_t$ to $\hat{s}_{t+1}$ must correspond to a realizable sensory change.
This contrasts with encoder-only latent matching objectives (e.g., JEPA-style losses; \citealp{assran2023self}) that align predicted latents with encoded future latents; as analyzed in~\citet{xing2025critiques}, purely latent-space objectives can admit degenerate solutions and provide a loose surrogate for observation-level consistency.
By decoding predictions back to the sensory domain, GLP directly constrains learned dynamics to be actionable and evidence-aligned.

\subsection{Design choices and parameterization}

\modelname implements the GLP architecture through three complementary modules (Figure~\ref{fig:model-arch}; full details in Supplementary Information \S1):
The \textbf{Vision Encoder}~$h$ adopts the vision tower of Qwen2.5-VL-7B-Instruct~\citep{bai2025qwen2}, mapping observations into spatially structured latent embeddings.
The \textbf{Autoregressive World-Model Backbone}~$f$ is based on the language model of Qwen2.5-VL-7B-Instruct. It predicts the next latent world state conditioned on the full history of latent states and language actions, modeling long-range temporal dependencies across extended rollouts. By adopting a pretrained LLM as the dynamics backbone, \modelname leverages multimodal representations and language-conditioned sequence modeling acquired through large-scale pretraining.
The \textbf{Video Diffusion Decoder}~$g$ is adapted from Wan2.1-T2V-14B~\citep{wan2025wan}. To reduce drift and compounding artifacts in extended rollouts, \modelname incorporates Causal Swin-DPM~\citep{feng2024matrix}, a chunk-wise causal denoising mechanism that uses a sliding temporal window holding two chunks at different noise levels, enabling smooth transitions across consecutive segments while a causal attention mask prevents information leakage from future actions (see Supplementary Information \S1.3 for the full mechanism).

Actions are specified as natural-language instructions at the video-chunk level. This interface targets language-conditioned visual rollout, simulating the qualitative consequences of alternative action choices such as moving toward an object, turning to a new scene, or manipulating an item; adaptation to diverse low-level continuous motor control interfaces is left to future work.
The discrepancy function $\mathrm{disc}(\cdot)$ is implemented with a flow-matching objective~\citep{lipman2022flow} to provide strong supervision for video generation.
The GLP architecture itself is not limited to these specific instantiations: each component can be independently enhanced to incorporate hierarchical embeddings, mixed discrete--continuous representations, or higher-order temporal dynamics, offering a flexible foundation for future extensions.

\section{Evaluation on world reasoning}\label{sec:results}

If the purpose of a world model is to serve as a simulator for reasoning by enabling an agent to predict the consequences of its actions, maintain a coherent picture of the world over extended horizons, and plan by comparing alternative futures, then evaluation must measure these functional capabilities directly.
Standard video-generation metrics such as FID and FVD quantify perceptual realism but are agnostic to whether the model can faithfully simulate the causal effect of an action, sustain consistency as rollouts lengthen, or produce simulations reliable enough to improve downstream decision-making.
A model that scores well on visual fidelity may still fail as a world model if its outputs do not respect the causal structure of the environment.
We therefore evaluate \modelname on the World Reasoning Arena (WR-Arena)~\citep{gao2026wrarena}, a benchmark designed around three complementary dimensions of world-model competence: \emph{action simulation fidelity} (does the model correctly simulate what happens when an action is taken?), \emph{long-horizon forecasting} (does the simulation remain coherent over many steps?), and \emph{simulative reasoning and planning} (can an agent use the model's predictions to make better decisions?).
These dimensions form a progressive hierarchy: faithful single-step simulation is necessary for multi-step coherence, which in turn is necessary for reliable planning.
Full benchmark specifications, evaluation code, and baselines are described in~\citet{gao2026wrarena} and Supplementary Information~\S4.

We compare \modelname against both open-source models: WAN~2.1 \& 2.2~\citep{wan2025wan}, Cosmos~1 \& 2~\citep{cosmos,nvidia-cosmos-predict2}, and V-JEPA~2~\citep{assran2025vjepa2selfsupervisedvideo}; and commercial systems: KLING~\citep{klingai-app-cn}, MiniMax (Hailuo)~\citep{hailuo-ai-minimax}, and Gen-3~\citep{runway-gen3-alpha}.
This set spans the design spectrum from pure video generators (WAN, KLING, MiniMax, Gen-3) through physics-oriented world models (Cosmos) to latent-only predictors (V-JEPA~2), providing a broad test of whether GLP's combination of latent dynamics and generative supervision offers a measurable advantage.
All reported scores are accompanied by 95\% confidence intervals: binomial success-rate metrics in the reasoning and planning dimension use analytic score intervals (Wilson for single proportions, Newcombe for differences), while the pipeline-based fidelity and forecast metrics use percentile bootstrap over evaluation instances.
Details of the evaluation are given in Supplementary Information~\S4.
Figure~\ref{fig:MainResults} summarizes performance across all three dimensions; we discuss each below.

\begin{figure}[!t]
  \centering
  \includegraphics[width=\linewidth]{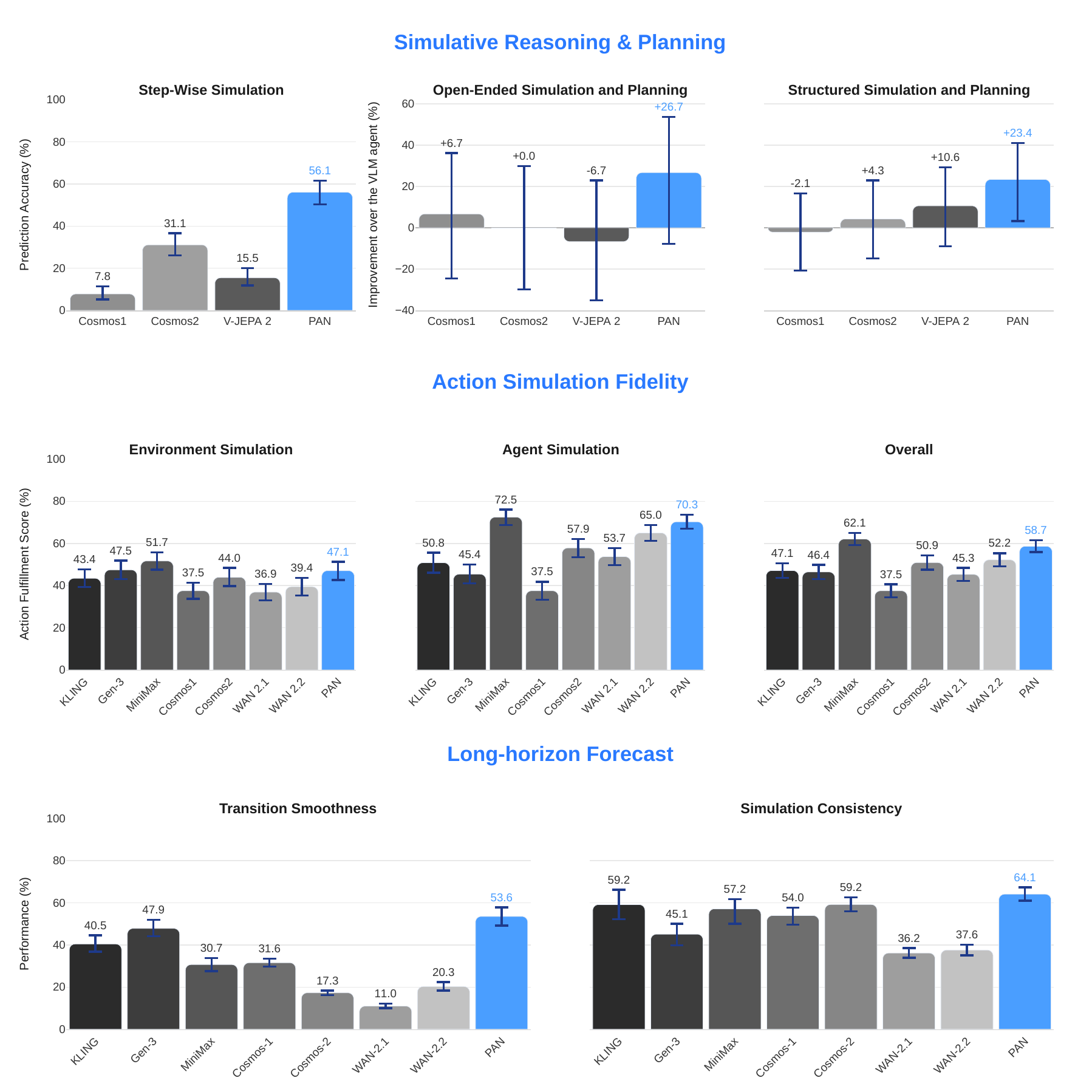}
  \caption{\textbf{Model performance on WR-Arena.} Scores for Action Simulation Fidelity, Long-Horizon Forecast, and Simulative Planning tasks. \modelname achieves the highest overall scores among open-source models and is competitive with commercial systems. Error bars represent 95\% confidence intervals ($n = 250$ for open-source models and $100$ for commercial systems). Additional details are provided in Supplementary Information \S4.}
  \label{fig:MainResults}
\end{figure}
\FloatBarrier

\paragraph{Action Simulation Fidelity.}
This dimension tests whether a model can correctly simulate the immediate visual consequences of a language-specified action, scored by a VLM-based judge on action faithfulness and precision.
\modelname achieves 70.3 on agent simulation and 47.1 on environment simulation, yielding an overall score of 58.7, the highest among all open-source models and comparable to the strongest commercial systems.
Qualitative rollouts across visually diverse environments, including stylized and non-photorealistic scenes, are shown in Extended Data Fig.~2.
The gap between agent and environment scores is informative: agent simulation asks the model to animate a controllable entity (e.g., ``the robot picks up the cup''), where the action target and its expected motion are relatively well-specified; environment simulation requires scene-level interventions (e.g., ``remove the chair,'' ``change the weather to rain'') that demand a broader understanding of how scenes transform holistically.
\modelname can nonetheless simulate uncommon or counterfactual scenarios, such as unusual weather events or atypical object behaviors, when prompted with corresponding action instructions (Extended Data Fig.~3).
The lower environment scores across all models indicate that scene-level causal reasoning remains harder than entity-level action execution.
Notably, video generation baselines, despite producing high-quality visual outputs, score substantially lower on action fidelity, particularly on multi-step sequences.
This gap is architectural: video generators produce complete sequences from fixed prompts and cannot condition each segment on the outcome of the preceding action, whereas \modelname performs step-wise, action-conditioned rollout in which each chunk is generated from an updated latent state.

\paragraph{Long-Horizon Forecast.}
This dimension measures whether simulation quality degrades gracefully as the rollout horizon extends.
\modelname achieves 53.6\% on Transition Smoothness and 64.1\% on Simulation Consistency, both the highest scores among all evaluated models including commercial systems.
Transition Smoothness quantifies motion continuity at chunk boundaries using dense optical-flow acceleration: a high score means that the generated motion evolves smoothly across the seam between consecutive chunks rather than exhibiting abrupt jumps.
\modelname's advantage of 5.7 points above the next-best model here is directly attributable to Causal Swin-DPM, which provides overlapping denoising context across adjacent chunks so that the later chunk ``sees'' partially denoised context from the earlier one.
Simulation Consistency tracks content alignment and style stability over the full length of the rollout, applying progressive penalties to later steps to emphasize the quality of predictions where drift typically becomes most severe.
\modelname's lead on this metric is narrower but consistent, indicating that the GLP architecture's latent state carries enough persistent information to prevent the rapid collapse observed in standalone video generators, where visual quality can deteriorate substantially within 5--8 rollout steps.
The long-horizon analysis in Supplementary Information~\S6.1 shows that PAN's consistency degrades gradually over 9 rollout steps while standalone generators deteriorate rapidly from their initial consistency. The progressive ablation (Supplementary Information~\S6.2--6.3) further decomposes this stability: the Causal Swin-DPM training substantially improves long-horizon simulation relative to the base model, while the VLM-based backbone contributes additional gains in agent dynamics, transition smoothness, and simulation consistency.

\begin{figure}[t]
  \centering
    \includegraphics[width=1\linewidth]{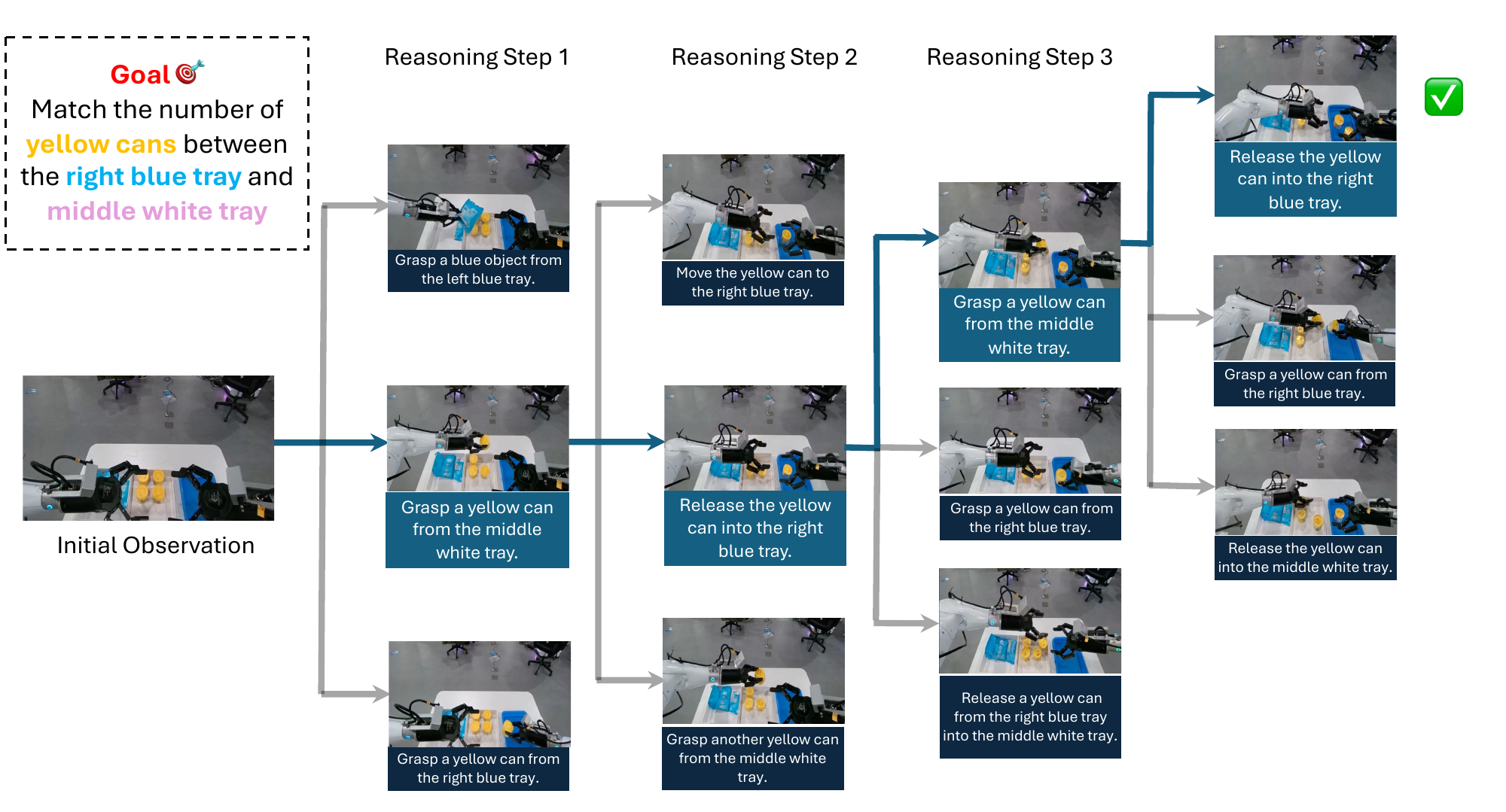}
    \caption{\textbf{Tree-structured simulative planning with PAN.} Given an initial state and a goal specification, a VLM proposes candidate actions at each step. The world model simulates each candidate's outcome as a ``thought experiment'', and the VLM selects the action whose predicted state shows maximal goal progress. Gray arrows show explored alternatives; blue arrows trace the selected trajectory.}
    \label{fig:planning_demo}
\end{figure}

\paragraph{Simulative Reasoning and Planning.}
The ultimate test of a world model is whether its simulations are reliable enough to improve an agent's decision-making.
This dimension evaluates exactly that: a VLM planning agent~\citep{OpenAI2025o3o4minisystemcard} proposes candidate actions, the world model simulates their consequences, and the agent selects the action whose predicted outcome best advances a specified goal (Figure~\ref{fig:planning_demo}).
On \emph{Step-Wise Simulation}, which measures atomic causal prediction in robotic manipulation scenarios, \modelname achieves 56.1\%, the highest among open-source baselines.
This score reflects the model's ability to predict, for a single action, the correct resulting spatial and physical configuration; errors at this level propagate multiplicatively through multi-step planning, so even modest improvements in single-step accuracy yield meaningful downstream gains.

On \emph{Open-Ended Planning} and \emph{Structured Planning}, integrating \modelname with the VLM agent increases task success by 26.7\% and 23.4\%, respectively, relative to the agent acting without any world model.
These are substantial margins that confirm the simulations are not merely visually plausible but causally informative. The agent makes better decisions because \modelname's predicted outcomes accurately distinguish actions that advance the goal from those that do not.
The contrast with baseline world models is striking: when the same agent is paired with other models, performance is inconsistent, with some baselines improving success on certain tasks but actively degrading it on others, suggesting that their simulations are sometimes realistic enough to appear helpful but causally inaccurate enough to mislead the planner.
This pattern underscores a key finding: visual realism and causal reliability are distinct properties, and only models that achieve both can consistently serve as planning tools.
\modelname's advantage in this dimension follows directly from the GLP design: because latent transitions are supervised through observation-space reconstruction, the model is trained not just to produce visually plausible futures but to produce futures that correctly reflect the causal consequences of the conditioned action.

\section{Discussion}\label{sec:discussion}

The central premise of this work, and of the broader program it belongs to, is that a world model should be built for and evaluated not as a media generator, but as a \emph{simulator for reasoning}.
As argued in~\citet{xing2025critiques}, the primary purpose of a world model is to enable an agent to perform hypothetical thinking, where the agent simulates the consequences of alternative actions and extracts the best course of action from among them.
One might further ask: where do proposals or actuations of actions come from, and should the world model be responsible for those as well?

The PAN framework argues for a structured treatment grounded in the agent--environment factorization~\citep{xing2025critiques}. The joint distribution over an interaction trajectory decomposes into two factors: the \emph{universe factor} $p_\mu(s' \mid s, a)$, which predicts what \emph{will happen} given a state and an action, and a functionally separate \emph{agent factor} $p_\pi(a \mid s, g)$, which decides what \emph{to do} given a state and a goal.
A world model (WM) is a learned approximation of the former; an agent model (AM) is a learned realization of the latter~\citep{xing2026critique}.
Crucially, these two models require different training signals: the world model improves by minimizing prediction error against observed state transitions, whereas the agent model improves through goal-directed reward feedback.
Collapsing both into a single model, as several recent proposals advocate~\citep{ye2026dreamzero,li2026functional_taxonomy_world_models,nvidia2026cosmos3}, conflates fidelity-driven next-state prediction with reward-driven action selection, undermining the reliability of both: the world model's predictions become biased toward rewarding outcomes rather than faithful ones, and the agent's decisions lose the ability to compare alternative futures against a neutral simulator.
PAN's design reflects this principle: it is a world model that predicts observation-grounded future states, while the planning agent that queries it~\citep{OpenAI2025o3o4minisystemcard} remains a separate module.
This separation has been validated in prior work on LLM-based simulative reasoning~\citep{hao2023reasoning,deng2025general}, where maintaining a distinct world model for planning consistently outperforms reactive baselines and monolithic approaches.

The GLP architecture embodies three principles about how such a simulator should be built.
First, latent dynamics, when carried by stateful, symbolic, and action-conditionable representations, are necessary for scalable, long-horizon reasoning; raw pixel sequences lack the abstraction and structure for planning, and fixed-size continuous embeddings scale poorly with world complexity (Theorem~1; \citealp{xing2025critiques}).
Second, a proper backbone must be adopted to coherently process both discrete and continuous representations: GLP's hierarchical structure assigns compositional and long-horizon reasoning to the LLM backbone over symbolic tokens and fine-grained perceptual dynamics to a diffusion-based predictor, absorbing the signal variability that makes single-modality architectures brittle.
Third, those dynamics must be \emph{generatively supervised}: each predicted latent transition must be decoded back to the sensory domain and compared against observable data, so that the representation is continuously anchored to reality.
This closed-loop design avoids the degeneracies of purely latent-space objectives, namely representation collapse (Proposition~1; \citealp{xing2025critiques}), information loss through unconstrained abstraction, and the brittleness of heuristic regularizers, while retaining the benefits of abstract, compositional state prediction.
PAN's ablation results support this design: removing the VLM backbone while retaining the generative decoder preserves short-horizon fidelity but degrades long-horizon consistency and agent dynamics, and in extended rollouts the full GLP system sustains coherent simulation where standalone generators degrade rapidly (Supplementary Information~\S6).

A persistent objection to generative world models is that reconstructing future observations in full detail is intractable and wasteful~\citep{lecun2022path}.
We offer three counter-arguments grounded in both theory and the PAN system.
First, generative supervision does not require \emph{perfect} pixel-level reconstruction; it requires that the decoder provide a training signal anchored in observable reality.
As Theorem~2 of~\citet{xing2025critiques} shows, the generative loss lower-bounds the latent loss: any model that reconstructs well in observation space necessarily predicts well in latent space, but the converse does not hold.
Second, the practical cost of reconstruction can be managed through architectural design.
PAN's Causal Swin-DPM conditions each chunk on partially noised context from the preceding chunk, effectively dampening incidental pixel-level detail and focusing the decoder's capacity on semantically persistent structure.
This is one instance of a broader class of efficient generative techniques, including sparse attention mechanisms~\citep{zhang2025fast,zhang2026faster}, that make closed-loop supervision practical at scale.
Third, purely latent objectives carry hidden costs of their own: the regularizers required to prevent collapse (variance constraints, energy-based penalties, architectural asymmetries) introduce optimization complexity and hyperparameter sensitivity that are easy to underestimate.
The training-dynamics analysis (Supplementary Information \S6.4) confirms that generative supervision is stable in practice: PAN's loss trajectory shows smooth optimization across both training stages with a persistently decreasing slope, consistent with the view that closed-loop reconstruction is a tractable and well-behaved training objective at scale.

Evaluating open-domain world models is itself a frontier challenge.
Standard video-quality metrics (FID, FVD) measure perceptual realism but not the properties that matter for a simulator, including causal fidelity, action controllability, and long-horizon consistency.
WR-Arena~\citep{gao2026wrarena} takes a step toward evaluation protocols aligned with world-model function by measuring action simulation fidelity, long-horizon forecasting, and simulative planning, but much work remains: VLM-based judges may introduce systematic biases, human evaluation is subject to inter-annotator variability, and the field lacks shared evaluation platforms comparable to those available for language models.
Future benchmark development should prioritize horizon-dependent degradation curves, causal counterfactual tests, and standardized planning protocols that measure whether a model's simulations reliably improve downstream decision-making.

Several concrete limitations of the current PAN system point to broader open problems.
The language-level action interface is well-suited for high-level planning and counterfactual imagination but cannot express the continuous torque commands and closed-loop proprioceptive feedback required for fine-grained manipulation; extending GLP to support multi-resolution action representations is a natural next step.
Long-horizon drift (e.g., semantic inconsistency, object hallucination, and compounding artifacts over extended rollouts) remains an open challenge across all world model architectures, including commercial systems; the problem is likely to require both architectural advances (richer stateful representations, hierarchical planning) and larger-scale, more diverse training data.
PAN inherits visual priors from Wan~2.1, an open-source video generation model optimized for perceptual realism rather than physical correctness, which can manifest as visually plausible but causally incorrect interactions; training a world model from scratch with objectives, data, and architectures designed specifically for simulation is a promising direction.
More broadly, the path forward for world modeling lies in tighter integration of multimodal data, mixed discrete--continuous representations, and generative objectives that ground abstract reasoning in observable reality, which are the principles that GLP is designed to embody.

\section{Methods}

\subsection{Model implementation}

\modelname implements the GLP architecture with three tightly coupled modules.
The \textbf{vision encoder} adopts the vision tower of Qwen2.5-VL-7B-Instruct~\citep{bai2025qwen2}, a Vision Transformer optimized for high-resolution and sequential inputs. It partitions each video frame into $14{\times}14$ spatial patches, applies windowed self-attention with 2D rotary positional embeddings, and groups consecutive frames through 3D patch partitioning for native temporal coherence. The resulting spatially structured continuous embeddings serve as the latent world state $\hat{s}_t$ in the GLP formulation, retaining the spatial and temporal organization needed for downstream dynamics modeling.

The \textbf{autoregressive world-model backbone} is adapted from the language model of Qwen2.5-VL-7B-Instruct and constitutes the core of PAN's GLP instantiation. At each timestep, the backbone takes three inputs: the current visual state (encoded observation tokens), a natural-language action, and 256 learnable query embeddings. These are arranged in a multi-turn conversational format, alternating \texttt{<user>} turns (visual state $+$ action) and \texttt{<assistant>} turns (query embeddings)---that aligns with the pretrained VLM dialogue structure and naturally induces a temporal chain over the interaction history $(\hat{s}_1, a_1, \hat{s}_2, \ldots, \hat{s}_t, a_t)$. The output is a compact set of 256 continuous tokens representing the predicted next latent state $\hat{s}_{t+1}$. This bottleneck design is central to GLP: the 256-token representation forces the backbone to compress the dynamically relevant information from the full multimodal context, while the generative decoder (below) ensures that compression does not silently discard information needed for faithful simulation. By building on a pretrained LLM, the backbone inherits broad multimodal knowledge, language-conditioned sequence modeling, and compositional reasoning capacities that a continuous-only predictor would lack, namely supporting the symbolic and long-horizon reasoning that GLP requires of the dynamics module.

The \textbf{video diffusion decoder} is adapted from Wan2.1-T2V-14B~\citep{wan2025wan}, a 14B-parameter Diffusion Transformer. The predicted latent state $\hat{s}_{t+1}$ is linearly projected and fed to a newly added cross-attention stream in each transformer block, with a zero-initialized output projection~\citep{zhang2023adding} to ensure stable training. In parallel, the action text is encoded with umT5~\citep{chung2023unimax} and provided to the original text cross-attention pathway; the two streams are summed, enabling the decoder to integrate global state context with action-specific visual changes. This decoder serves as the generative supervision engine of GLP: its reconstruction loss anchors every latent transition to observable reality. To reduce drift and compounding artifacts across extended rollouts, PAN incorporates Causal Swin-DPM~\citep{feng2024matrix}, a chunk-wise causal denoising mechanism described below.

\subsection{Causal Swin-DPM}

Causal Swin-DPM is a chunk-wise causal extension of the Shift-Window Denoising Process Model~\citep{feng2024matrix}, designed to make generative reconstruction practical over long rollouts. The mechanism operates within a sliding temporal window holding two chunks at different noise levels: given $K = 1000$ total denoising steps, the earlier chunk begins at noise level $K/2$ while the later chunk begins at $K$. After $K/2$ steps the earlier chunk is fully denoised and dequeued; a new Gaussian-noise chunk is enqueued. A chunk-wise causal attention mask ensures the later chunk attends only to the earlier one, preventing information leakage from future actions. Crucially, by conditioning on \emph{partially noised} rather than fully denoised context, the mechanism dampens incidental pixel-level detail from the previous chunk and encourages the decoder to rely on semantically persistent information (scene structure, object presence, motion direction). This separation of local perceptual variation from higher-level continuity signals addresses a key practical concern about generative supervision, namely that the decoder would be forced to reproduce every fine-grained detail, while retaining the observation-space training signal that GLP requires. During training, the noise-level difference of $K/2$ is enforced by subsampling $k$ from $[0, 0.5]$ for the first chunk and assigning $k + 0.5$ to the second. Full training procedure and conditioning-frame noise augmentation are described in Supplementary Information \S1.3--1.4.

\subsection{Training}

Training follows a two-stage pipeline (see Supplementary Information \S2 for full details). In \textbf{Stage~1 (Module-wise training)}, the video diffusion decoder is adapted from Wan2.1-T2V-14B into the Causal Swin-DPM architecture while the vision encoder and backbone remain frozen at their pretrained Qwen2.5-VL initialization. Training uses BFloat16 precision, the AdamW optimizer~\citep{loshchilov2017decoupled} with learning rate $1 \times 10^{-5}$, cosine decay with 5\% warm-up, and gradient clipping at 0.05. The model is trained for 5 epochs on 960 NVIDIA H200 GPUs using Hybrid Sharded Data Parallelism with activation checkpointing.
In \textbf{Stage~2 (Joint training)}, the full system is trained end-to-end under the GLP generative supervision objective, coupling latent prediction with observation-space reconstruction via the flow-matching loss. The vision-language model is frozen; only the query embeddings and the video diffusion decoder are trained. The world model history is restricted to the most recent 10 rounds to align with the Qwen2.5-VL context window. Training uses the same optimizer settings as Stage~1, with early stopping after 1 epoch based on validation convergence. Sequence parallelism~\citep{li2021sequence} with the Ulysses method~\citep{jacobs2023deepspeed} is applied to manage memory for the large sequence lengths in both the VLM and DiT.

\subsection{Training data}

The dataset is drawn from publicly accessible video sources spanning everyday human activities, human--object and human--environment interactions, natural and urban scenes, embodied navigation, and multi-agent scenarios. Raw long-form videos are segmented using a dynamic shot-boundary detection strategy~\citep{chen2024panda70m} and filtered through a multi-stage pipeline combining rule-based motion and quality filters, pretrained detectors for aesthetic quality and obstructive text, and a custom-trained VLM filter for content categories unsuitable for world modeling. The filtering pipeline retains 16.1\% of original clips. Dense captions focusing on temporal dynamics rather than static scene attributes are generated using a vision--language model, yielding 8 million video--text pairs corresponding to approximately 4,500 hours of video. Full data pipeline details and filtering statistics are provided in Supplementary Information \S3.

\subsection{Evaluation protocol}

We adopt the three-dimensional evaluation framework from WR-Arena~\citep{gao2026wrarena}: \textbf{Action Simulation Fidelity} evaluates how faithfully the model simulates language-specified actions and their causal consequences, scored by a VLM-based judge following the protocol of~\citet{worldmodelbench}; \textbf{Long-horizon Forecast} measures temporal continuity (via dense optical flow) and content consistency over extended rollouts; and \textbf{Simulative Reasoning and Planning} evaluates whether the world model can serve as an internal simulator enabling a VLM agent~\citep{OpenAI2025o3o4minisystemcard} to reason about actions and plan toward goals, using robotic manipulation scenarios from Agibot~\citep{bu2025agibot} and tabletop environments from Language Table~\citep{lynch2023interactive}. Full protocol details, metric definitions, and baseline descriptions are provided in~\citet{gao2026wrarena} and Supplementary Information \S4.

We compare \modelname with open-source models including WAN 2.1 \& 2.2~\citep{wan2025wan}, Cosmos 1 \& 2~\citep{cosmos,nvidia-cosmos-predict2}, and V-JEPA 2~\citep{assran2025vjepa2selfsupervisedvideo}, and closed-source models including KLING~\citep{klingai-app-cn}, MiniMax (Hailuo)~\citep{hailuo-ai-minimax}, and Gen-3~\citep{runway-gen3-alpha}. Baseline descriptions are in Supplementary Information \S4.2.

\subsection{Inference}

During inference, \modelname performs autoregressive multi-step simulation: given an initial observation $o_1$, it estimates $\hat{s}_1 = h(o_1)$ and sequentially predicts future states and reconstructs observations. To improve long-term consistency, predicted states are augmented with encoder outputs of the corresponding reconstructed observations, forming enhanced states $\hat{s}'_k = [\hat{s}_k, h(\hat{o}_k)]$. Classifier-free guidance~\citep{ho2022classifier} with a guidance scale of 4 is applied during decoding. Inference acceleration techniques, including sequence parallelism and quantized attention via SageAttention2++~\citep{zhang2025sageattention2++} (achieving 30.3\% speedup with minimal quality impact), are described in Supplementary Information \S2.3.

\subsection{Data availability}

PAN is trained on 8 million video--text pairs (${\sim}$4{,}500 hours of video) drawn from publicly accessible sources.
Comprehensive statistics and details of the data-construction methodology are presented in Supplementary Information, \S3. 



\clearpage
\section*{Author List}

\noindent
Zihan Liu$^{\dagger}$, Yi Gu$^{\dagger}$, Mingkai Deng$^{\dagger}$,
Guangyi Liu$^{\dagger}$, Zeyu Feng, Qiyue Gao, Yiyan Hu,
Benhao Huang, Yichi Yang, Kun Zhou, Jiannan Xiang,
Zhiting Hu, Zhengzhong Liu, and Eric P. Xing.


\noindent
$^{\dagger}$ Co-first authors


\section*{Author Contributions}

Zihan L., Y.G., M.D., G.L. and J.X. led the development and evaluation of the system. Z.F. and B.H. developed the data engineering pipeline. Q.G. developed the evaluation framework. Y.Y. contributed to model architecture design. Y.H. and K.Z. contributed to software engineering. Zihan L., Y.G., M.D., G.L. and E.P.X. wrote the paper. Z.H., Zhengzhong L., and E.P.X. conceived the project; Z.H. contributed to early model architecture design; Zhengzhong L. managed the research; E.P.X. designed and led the project. All authors reviewed and approved the final manuscript.

\section*{Corresponding Author}
Correspondence and requests for materials should be addressed to Eric P. Xing (eric.xing@mbzuai.ac.ae).

\section*{Competing Interests}
The authors declare no competing interests.

\section*{Additional Information}
Supplementary Information is available for this paper.

\section*{Use of Large Language Models}
LLM tools were used to provide writing advice and assist editing during the preparation of this manuscript.

\clearpage

\section*{Extended Data}\label{sec:extended-data}

\subsection*{Extended Data Fig.\ 1: Interactive long-horizon simulation}
Multi-step interactive simulation in which the model generates consecutive video segments conditioned on a sequence of natural-language action instructions, starting from a single initial frame. Examples span kitchen manipulation and autonomous driving scenarios over 10+ action steps.

\begin{figure}[H]
\centering
\includegraphics[width=0.85\textwidth]{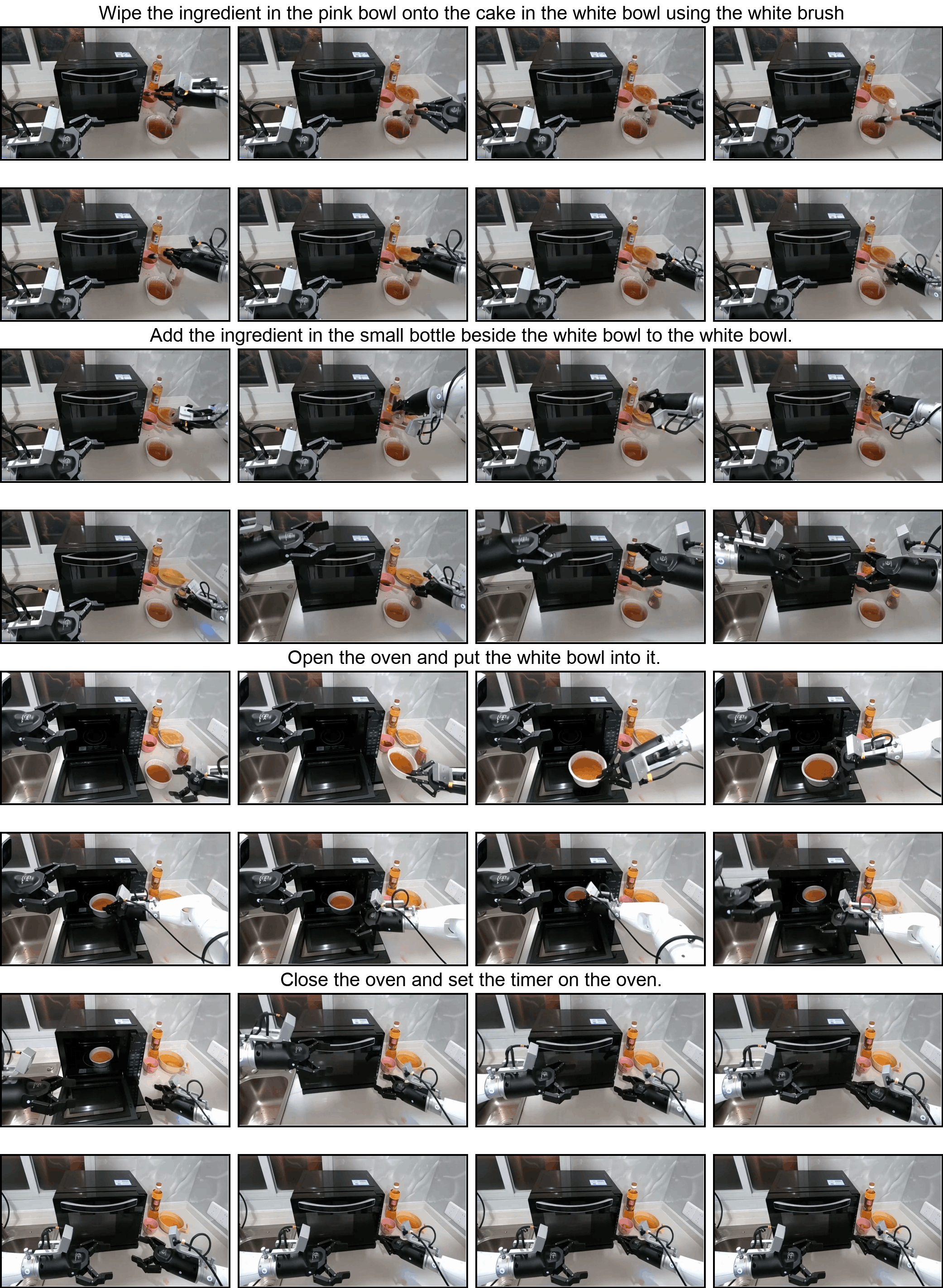}
\end{figure}

\begin{figure}[H]
\centering
\includegraphics[width=0.85\textwidth]{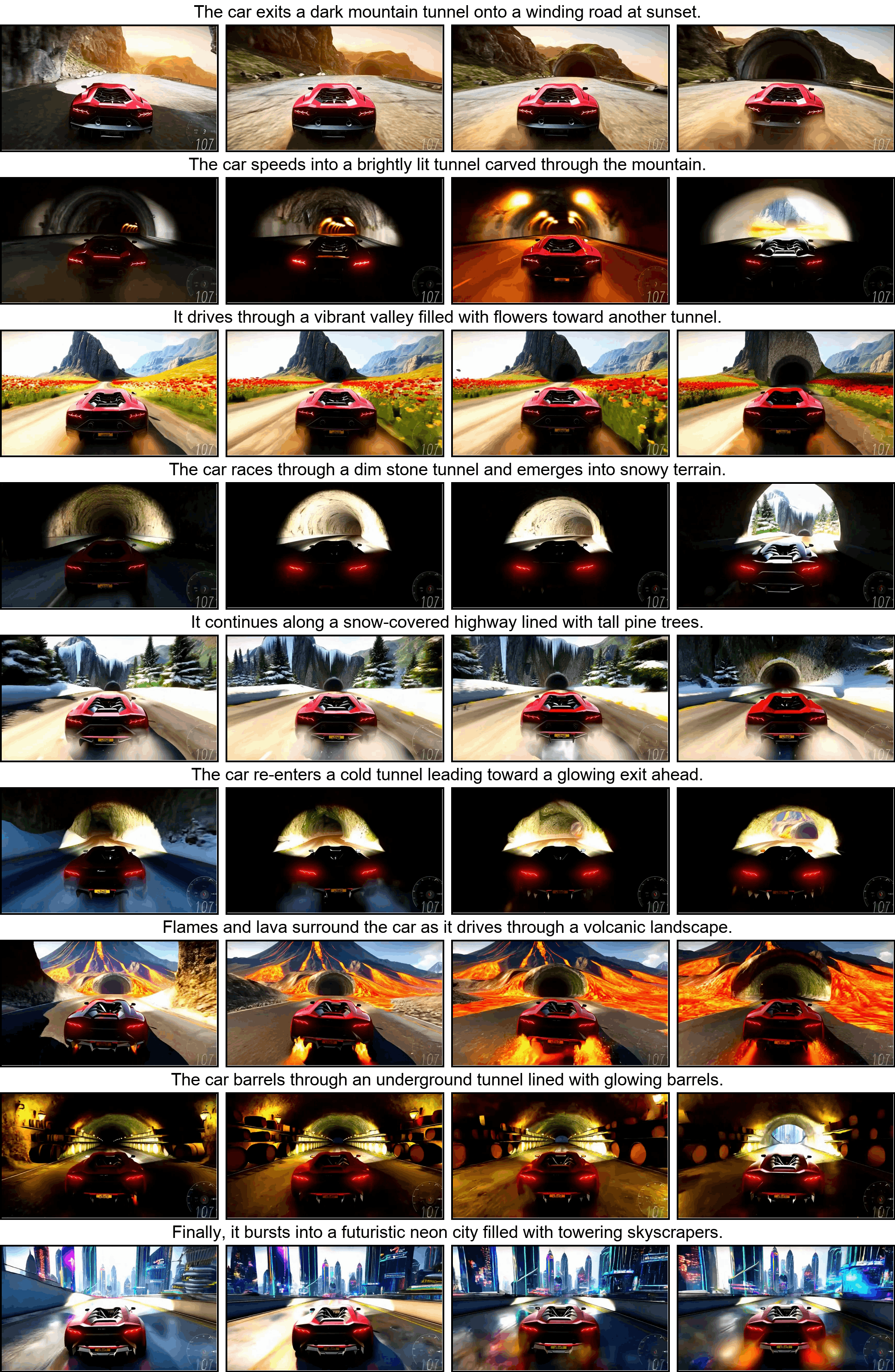}
\end{figure}

\pagebreak

\subsection*{Extended Data Fig.\ 2: Diverse action and world simulation}
The model follows diverse action instructions across visually distinct environments, including stylized and non-photorealistic scenes. The same action types (e.g., ``move forward,'' ``turn left'') produce domain-appropriate visual outcomes in each setting.

\begin{figure}[H]
\centering
\includegraphics[width=0.85\textwidth]{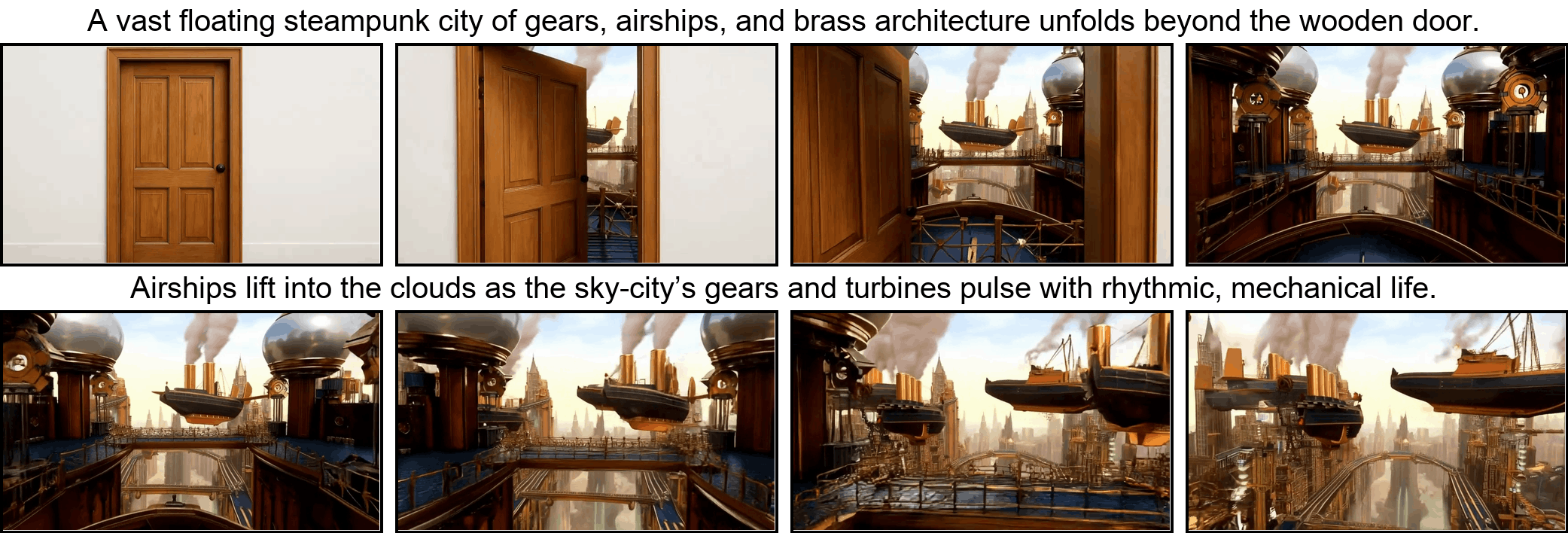}
\vspace{0.1cm}
\includegraphics[width=0.85\textwidth]{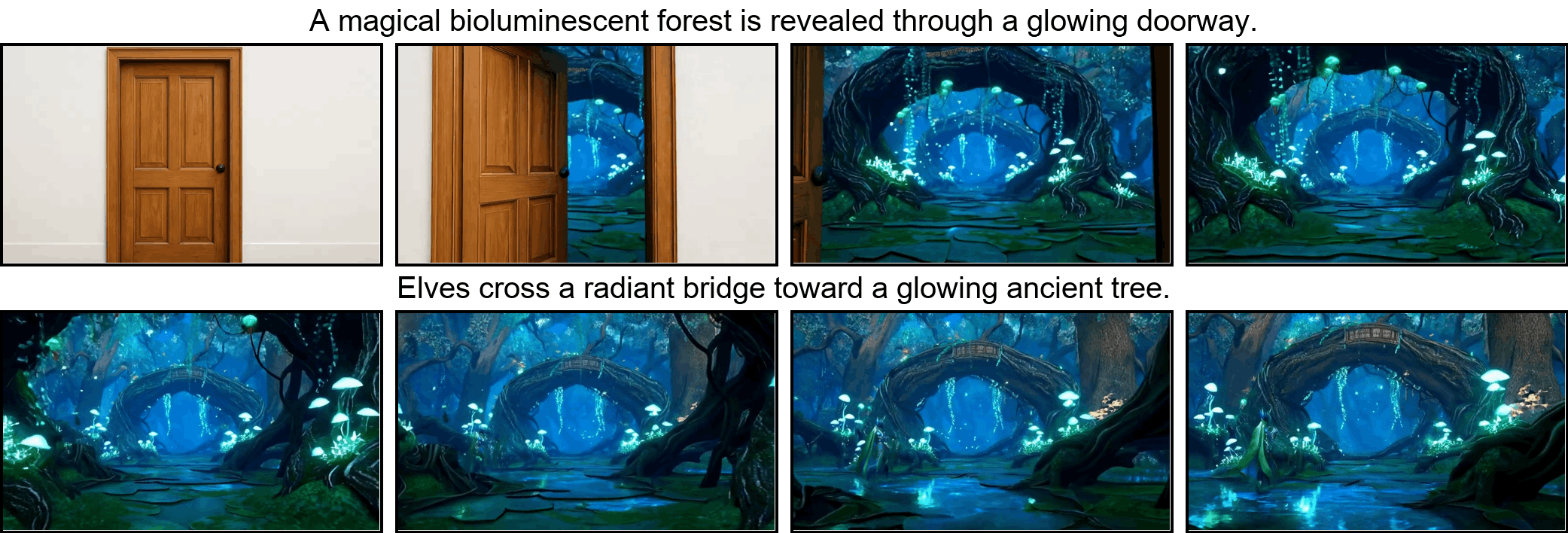}
\vspace{0.1cm}
\includegraphics[width=0.85\textwidth]{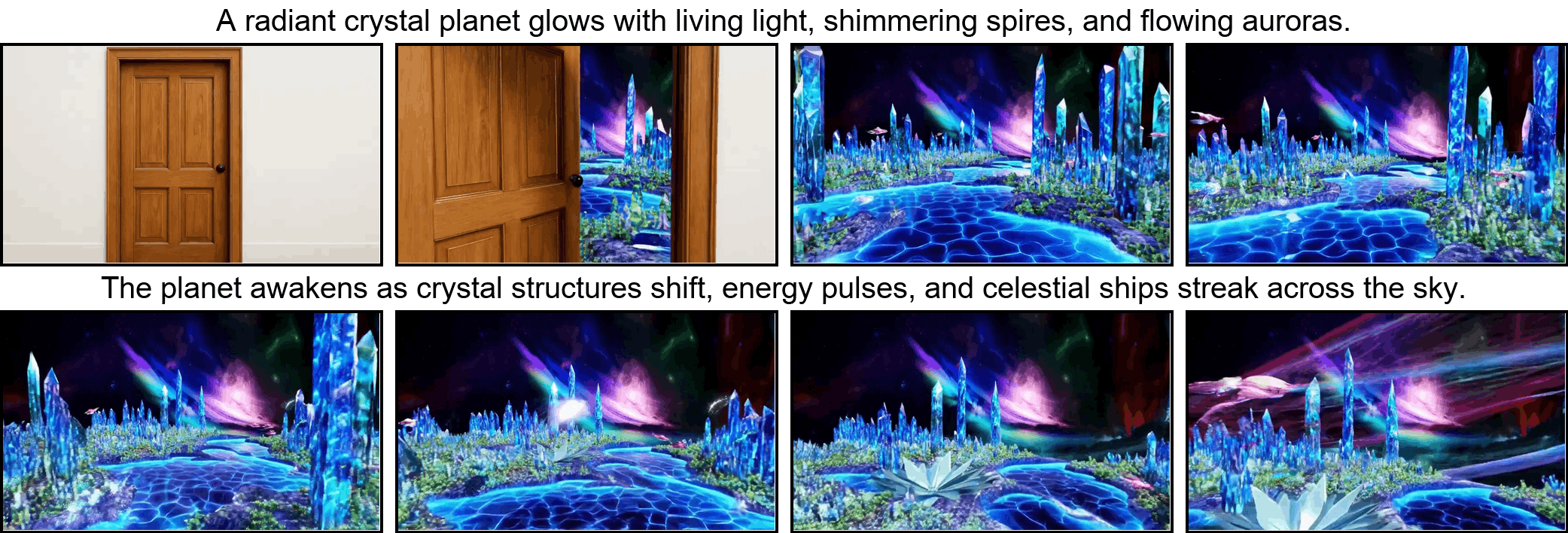}
\vspace{0.1cm}
\includegraphics[width=0.85\textwidth]{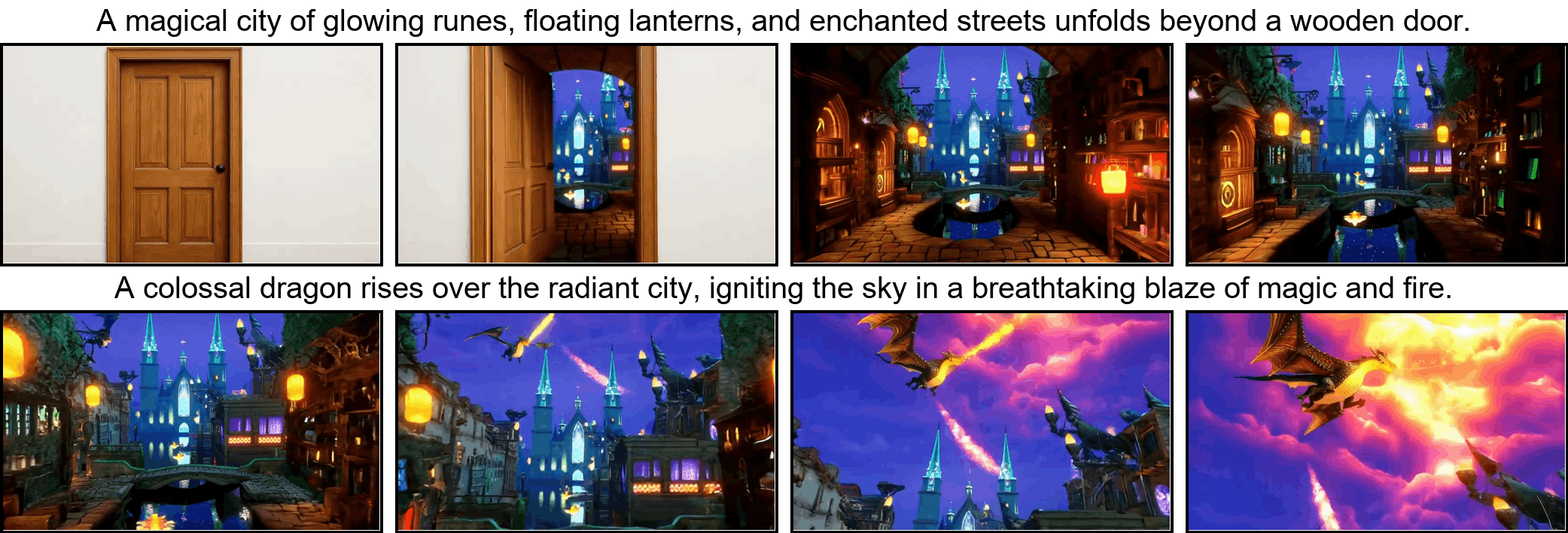}
\end{figure}

\subsection*{Extended Data Fig.\ 3: Rare case simulation}
The model generates uncommon or counterfactual scenarios (e.g., unusual weather events, mechanical failures, or atypical object behaviors) when prompted with corresponding action instructions.

\begin{figure}[H]
\centering
\includegraphics[width=1\textwidth]{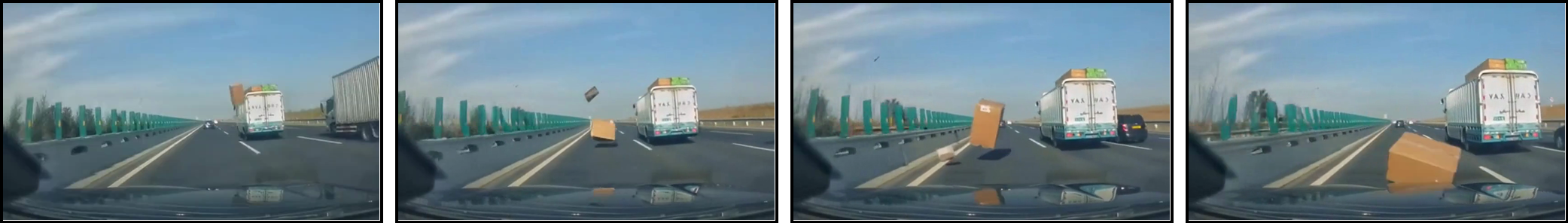}
\includegraphics[width=1\textwidth]{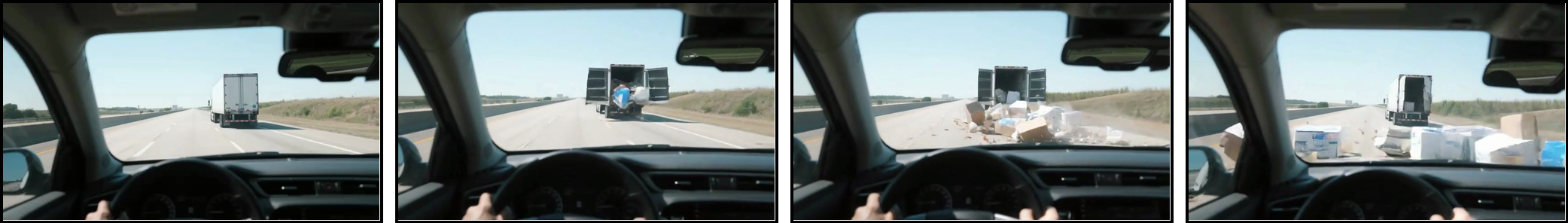}
\includegraphics[width=1\textwidth]{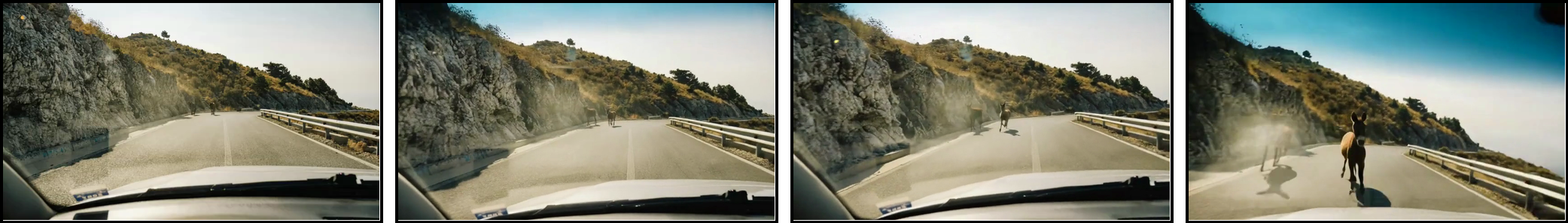}
\includegraphics[width=1\textwidth]{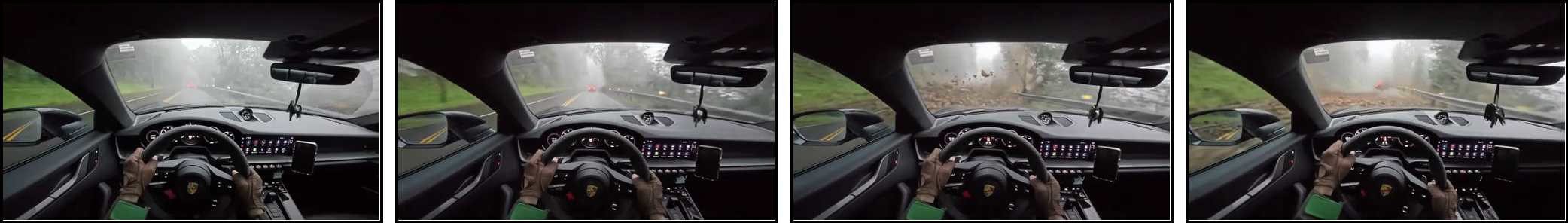}
\end{figure}

\clearpage
\addtocontents{toc}{\protect\setcounter{tocdepth}{3}}
\setcounter{section}{0}
\setcounter{figure}{0}
\setcounter{table}{0}
\setcounter{equation}{0}
\renewcommand{\thesection}{\arabic{section}}
\renewcommand{\thefigure}{\arabic{figure}}
\renewcommand{\thetable}{\arabic{table}}
\renewcommand{\theequation}{\arabic{equation}}
\renewcommand{\thefigure}{S\arabic{figure}}
\renewcommand{\thetable}{S\arabic{table}}
\renewcommand{\theequation}{S\arabic{equation}}

\renewcommand{\theHsection}{SI.\arabic{section}}
\renewcommand{\theHsubsection}{SI.\arabic{section}.\arabic{subsection}}
\renewcommand{\theHsubsubsection}{SI.\arabic{section}.\arabic{subsection}.\arabic{subsubsection}}
\renewcommand{\theHfigure}{SI.\arabic{figure}}
\renewcommand{\theHtable}{SI.\arabic{table}}
\renewcommand{\theHequation}{SI.\arabic{equation}}

\begin{center}
{\LARGE\bfseries Supplementary Information}\\[0.5cm]

\end{center}
\tableofcontents

\newpage

\section{Model Architecture Details}\label{si:arch}

This section provides full architectural details for each component of \modelname. A summary is given in the main text; here we describe the design choices, conditioning mechanisms, and the Causal Swin-DPM denoising process in full.

\subsection{Vision Encoder}\label{si:vision-encoder}

The vision encoder $h$ provides the perceptual representation used by the PAN predictive backbone. It is adopted from the vision tower of Qwen2.5-VL-7B-Instruct~\citep{bai2025qwen2}, a Vision Transformer (ViT) optimized for high-resolution and sequential inputs. The encoder partitions each video frame into $14{\times}14$ spatial patches, applies windowed self-attention~\citep{vaswani2017attention} for computational efficiency, and encodes the resulting tokens with 2D rotary positional embeddings~\citep{su2024roformer} that preserve spatial structure. For video streams, consecutive frames are grouped through 3D patch partitioning, allowing the encoder to represent short-term motion and temporal coherence natively within its token sequence.

The encoded visual tokens form the perceptual representations of the current world state, which retain spatial and temporal organization, providing structured visual features for the autoregressive backbone instead of requiring direct prediction over raw pixels. Within the GLP framework, the encoder fulfills the role of $h$, transforming observations into latent representations used by the predictive backbone $f$.

\subsection{Autoregressive World Model Backbone}\label{si:backbone}

The autoregressive world model backbone $f$ predicts a compact latent representation for the next video segment conditioned on the sequence of previous visual states and natural-language actions. Unlike the simplified GLP notation in the main text, the implemented backbone is non-Markovian: its input contains the interaction history $(\hat{s}_1, a_1, \hat{s}_2, \ldots, \hat{s}_t, a_t)$ within the context window. It rolls forward the state of the world step by step in latent form, ensuring that each predicted state is informed by both historical context and the current action.

The backbone is adapted from the language model of Qwen2.5-VL-7B-Instruct, which provides a pretrained multimodal sequence model and a dialogue-style interface for interleaving visual states and language actions. Following the GLP paradigm, the backbone corresponds to the predictive module $f$, which evolves the latent state step by step, ensuring that each transition captures the causal consequences of the corresponding action.

\subsubsection{Observation and Action Conditioning}\label{si:obs-action}

To seamlessly integrate observation encoding and action conditioning within the pretrained backbone, \modelname organizes each simulation step as a structured multi-turn conversation. A system prompt defines the task; each action is issued as a user turn in natural language; the visual state is attached as an inline image; and the backbone's response takes the form of latent world-state tokens:

\begin{tcolorbox}[colback=gray!5!white, colframe=gray!75!black, title={Input template for the VLM in the PAN autoregressive backbone}]
\ttfamily
<|user|> <image state (video state 1)> <action 1>

<|assistant|> <query embedding * 256>

<|user|> <video state 2> <action 2>

<|assistant|> <query embedding * 256>

......

<|user|> <video state n-1> <action n-1>

<|assistant|> <query embedding * 256>
\end{tcolorbox}

This conversational format leverages the pretrained dialogue structure of Qwen2.5-VL-7B-Instruct, allowing the backbone to process interleaved visual and textual inputs without architectural modification. Each round of the conversation corresponds to one simulation step $t$, with the user turn providing the current observation $o_t$ and action $a_t$, and the assistant turn producing the next latent world state $\hat{s}_{t+1}$.

\subsubsection{Latent World Prediction}\label{si:latent-prediction}

At each time step, the backbone generates the next latent world state as a fixed-length sequence of 256 continuous tokens. These tokens are produced by passing 256 learnable query embeddings through the backbone alongside the observation and action context. The backbone processes the full context and outputs 256 continuous vectors in a shared multimodal embedding space, capturing both visual and semantic aspects of the predicted world state. This compact latent representation is then passed to the video diffusion decoder for observation reconstruction.

\subsection{Video Diffusion Decoder}\label{si:decoder}

The video diffusion decoder $g$ is adapted from Wan2.1-T2V-14B~\citep{wan2025wan}, a 14-billion-parameter Diffusion Transformer (DiT) with 4x temporal downsampling. The decoder generates the next observation video $\hat{o}_{t+1}$ conditioned on the latent world state $\hat{s}_{t+1}$ and the action $a_t$. To support long-horizon interactive simulation, the decoder is extended with the Causal Swin-DPM mechanism described below.

\subsubsection{Flow Matching Objective}\label{si:flow-matching}

The discrepancy function $\mathrm{disc}(\hat{o}_{t+1}, o_{t+1})$ in the GLP objective is implemented as a flow-matching loss. Specifically, we adopt the Rectified Flow formulation~\citep{lipman2022flow,liu2022flow,zhoudenoising}, which defines a linear interpolation path between a noise sample and the target observation:
\begin{align}
x_k = kx_1 + (1 - k)x_0.
\end{align}
The model is trained to predict the ground-truth velocity $v_k$, defined as:
\begin{align}
v_k = \frac{\mathrm{d}x_k}{\mathrm{d}k} = x_1 - x_0.
\end{align}
We use 1000 denoising steps in training, with $k$ sampled from 1000 discrete values within the interval $[0, 1]$ using a shifted denoising step schedule~\citep{esser2024scaling}. To accommodate the sliding window mechanism in our Causal Swin-DPM model, we incorporate specific modifications to the loss function, such as restricting the timestep sampling range, as described in \S\ref{si:swdpm}.

\subsubsection{Conditioning on Actions and World State}\label{si:conditioning}

The decoder conditions on two sources of information: the latent world state $\hat{s}_{t+1}$ from the backbone and the action text $a_t$. The latent world state is first linearly projected to the conditioning dimension of the decoder, and then fed to a newly added cross-attention stream in each attention block. The output of this stream passes through a second linear projection that is zero-initialized to ensure stable training, following the strategy proposed in~\citet{zhang2023adding}. In parallel, the action text is encoded with umT5~\citep{chung2023unimax} and provided to the original text cross-attention pathway. Within each block, the post-projection output of the world-state stream is summed with the text-conditioned output, enabling the decoder to integrate global state context with action-specific visual changes while maintaining short-term local consistency.

The DiT block architecture is illustrated in Supplementary Figure~\ref{fig:dit-arch}.

\begin{figure}[H]
\centering
\includegraphics[width=0.5\linewidth]{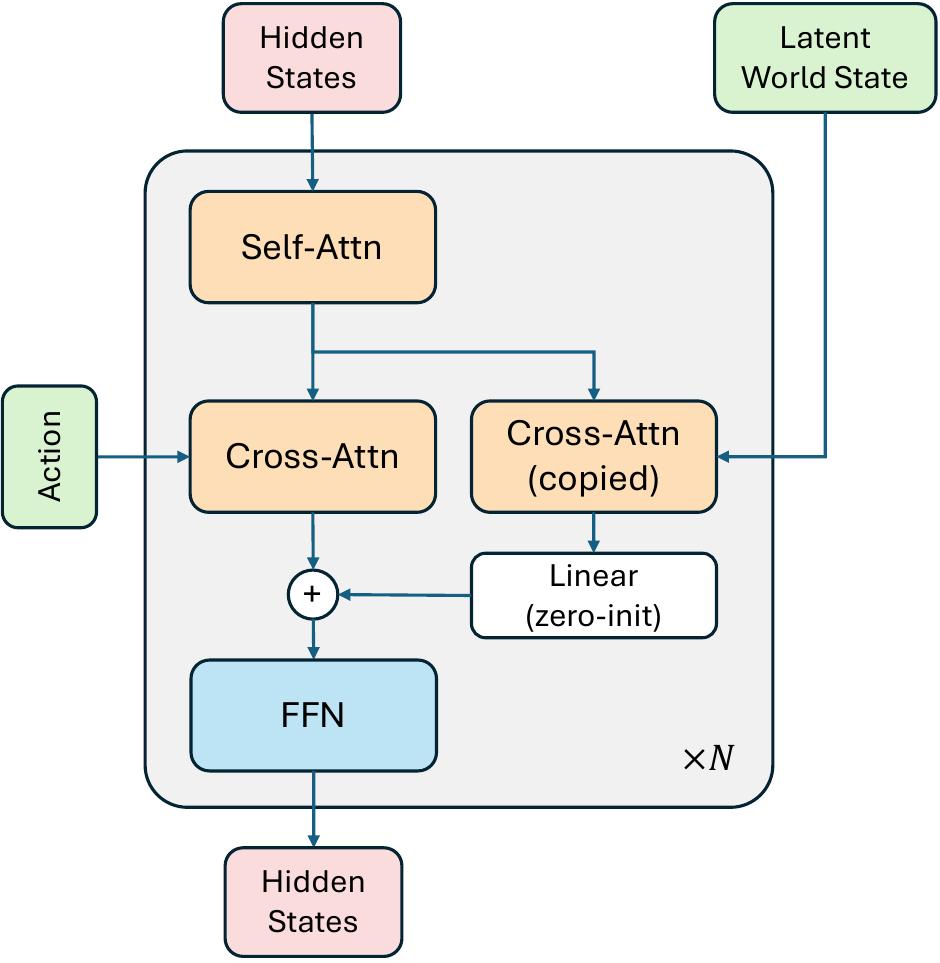}
\caption{Transformer block of the video diffusion decoder in \modelname. The latent world state conditions the decoder through a dedicated cross-attention stream with zero-initialized output projection. Action text is encoded via umT5 and provided through the standard text cross-attention pathway.}
\label{fig:dit-arch}
\end{figure}

\subsubsection{Causal Swin-DPM}\label{si:swdpm}

Traditional video diffusion models~\citep{wan2025wan,kong2024hunyuanvideo,veo2024,videoworldsimulators2024} are typically designed for single-shot generation. A straightforward way to extend them for sequential world simulation is to use the last frame from the last predicted observation as the condition for simulating the next one, repeating this process step by step. However, this naive extension causes two significant problems in multi-chunk autoregressive rollout:

\begin{enumerate}[leftmargin=*]
\item \textbf{Local inconsistency between adjacent chunks.} When the model conditions only on a single final frame rather than the entire denoising trajectory of the preceding observation, the transition between two video chunks may suffer from abrupt changes in motion or appearance. These discontinuities disrupt short-term coherence and make the simulation feel visually disconnected.

\item \textbf{Rapid quality degradation over extended rollouts.} Conditioning on only the last frame means that any small artifacts, motion drift, or misalignments introduced in one predicted observation are directly passed to the next. As the number of steps increases, these imperfections can accumulate quickly, leading to noticeable deterioration in visual fidelity and stability.
\end{enumerate}

Shift-Window DPM generates long videos by denoising temporal windows sequentially rather than producing the full sequence in a single pass. PAN modifies this idea for action-conditioned autoregressive rollout by imposing a chunk-wise causal attention mask, so that the current chunk can use past denoising context while preventing access to future action-conditioned content.

To address the above sources of discontinuity, we introduce Causal Swin-DPM, a chunk-wise causal extension of the Shift-Window Denoising Process Model~\citep{feng2024matrix}, as illustrated in Supplementary Figure~\ref{fig:swdpm}. This design is intended to improve local transitions and reduce visual discontinuities over multi-chunk rollouts while preserving autoregressive action conditioning.

Specifically, Causal Swin-DPM operates within a sliding temporal window that simultaneously holds two chunks of video frames at different noise levels. Suppose the total denoising steps are $K$ (i.e., 1000 for our model). At the beginning, video frames for the earlier chunk are at a noise level of $K/2$, while those for the later chunk are at a full noise level of $K$. After $K/2$ denoising steps, the frames from the earlier chunk are fully denoised and dequeued as the new video chunk. Concurrently, a new chunk of noisy frames initialized with Gaussian noise is enqueued at the end of the window. During the iterative denoising process, different chunks of video frames are conditioned on their corresponding natural language actions.

Beyond improving temporal continuity, Causal Swin-DPM also provides a structured way to condition on the recent generation history without requiring the next chunk to exactly copy all pixel-level details from the previous one. Rather than conditioning only on a fully denoised final frame, the mechanism uses partially noised representations of preceding chunks. This partially noised context reduces reliance on incidental pixel-level details from the previous chunk and encourages the decoder to use more persistent information such as scene structure, object presence, and motion direction. By coupling this context with the diffusion denoising trajectory, Causal Swin-DPM separates local perceptual variation from the higher-level continuity signals needed for chunk-to-chunk generation.

\begin{figure}[H]
\centering
\includegraphics[width=0.9\linewidth]{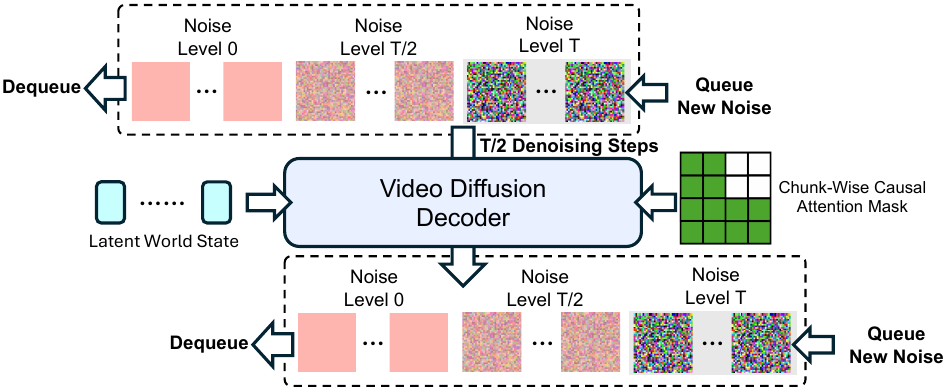}
\caption{Denoising process for Causal Swin-DPM. The sliding temporal window holds two chunks at different noise levels. A chunk-wise causal attention mask (green = visible, white = masked) prevents information leakage from future actions while allowing each chunk to attend to the denoising context of the previous chunk.}
\label{fig:swdpm}
\end{figure}

During training, we require two video frame chunks with a noise level difference of $K/2$, which corresponds to half of the denoising step sampling interval $[0, 1]$. To achieve this, we first subsample $k$ from $[0, 0.5]$ for the first chunk, and assign $k + 0.5$ to the second chunk. The only exception occurs when generating the first video chunk since it must be fully denoised from pure noise only conditioned on the given initial frame; we subsample $k$ from the full interval $[0, 1]$ in this case.

The ability of the sliding temporal window to see the context of the previous chunk of video frames allows for a reduction in one source of error accumulation by allowing each newly generated chunk to attend to partially denoised context from the previous chunk.

In typical inference settings, the natural language action for the next chunk will not be available until the current chunk is fully generated. To maintain real-time interactivity, the model employs a chunk-wise causal attention mask, as shown in Supplementary Figure~\ref{fig:swdpm}, where green cells indicate visible context and white cells represent masked positions corresponding to the future chunk with unseen actions. This masking ensures that the later video chunk can only attend to the previous one, preventing information leakage from future actions.

\subsubsection{Variational Autoencoder Padding}\label{si:vae}

We use the Variational Autoencoder (VAE) from Wan2.1 for video frame compression. Wan2.1-VAE is a 3D causal VAE~\citep{kingma2013auto,wu2025improved} that performs both temporal and spatial compression. It compresses $1 + T$ frames into $1 + T / 4$ latent features, and our model operates within the VAE latent space. For our video diffusion decoder, the window size is 21, which corresponds to 81 real video frames. During compression, the latent representation of each frame is conditioned on its preceding frames. Since our model uses a sliding window mechanism, the latent output at a given timestep may be influenced by a large number of preceding frames. To simulate this behavior during training, we randomly pad each 81-frame clip with 0 to 122 preceding frames. These additional frames are encoded by the VAE to provide proper temporal context, but are discarded afterward.

\subsubsection{Conditioning Frame and Noise Augmentation}\label{si:noise-aug}

The sliding window in Causal Swin-DPM contains one conditioning frame at the beginning, followed by two video chunks of size 10. The conditioning frame is taken as the last frame from the previously dequeued video chunk. When the window advances, it shifts by 10 latent frames, naturally moving the last frame of the current chunk to the front to serve as the new conditioning frame.

Empirically, we observe that using a fully denoised frame as the conditioning input can lead to error accumulation over long-horizon generation. To mitigate the problem, we apply noise augmentation to the conditioning frame. Specifically, we add Gaussian noise corresponding to a small fixed denoising step $k = 0.055$, which does not corrupt the conditioning frame excessively but still introduces sufficient stochasticity. During training, we do not compute loss on the conditioning frame. This noise augmentation is another implementation choice for aligning training with the partially noised context used in Causal Swin-DPM.

\section{Training Details}\label{si:training}

Training a general-purpose world model requires more than direct supervision from raw video or short-term observations. To achieve robust long-horizon reasoning and simulation, we adopt a divide-and-conquer strategy in which individual modules are first optimized for their own roles before being integrated into a unified system. This staged approach allows each component to develop stable and specialized capabilities, such as high-fidelity short-term state simulation or accurate modeling of long-term causal dynamics, before joint optimization aligns them for end-to-end performance.

\subsection{Stage 1: Module-Wise Training}\label{si:stage1}

Training individual modules separately allows each to develop stable and specialized capabilities before the complexities of full system integration. By optimizing components in isolation, we allow each to master its specific role and establish a strong initialization for later joint optimization.

For PAN, the vision encoder and the autoregressive backbone are built on the vision transformer and the joint vision-language transformer in Qwen2.5-VL-7B-Instruct, which has already undergone extensive pretraining and therefore requires no additional adaptation at this stage. The video diffusion decoder, on the other hand, is adapted from Wan2.1-T2V-14B, a non-causal video diffusion architecture originally designed for single-shot generation. In this stage, we adapt it into our Causal Swin-DPM architecture, which enables smooth temporal transitions and interactive action control over long-horizon world state simulation. We keep the linear weights in the decoder transformer block to be zero, so the model does not depend on the latent world state in this module-wise training stage.

Since we freeze both the Wan-VAE and the action text encoder in this stage, we precompute their latent features during data preprocessing. To reduce GPU memory usage, we adopt a combination of Data Parallelism~\citep{ben2019demystifying} and Fully Sharded Data Parallelism~\citep{zhao2023pytorch}. Specifically, FSDP is applied within each 8-GPU compute node, forming a Hybrid Sharded Data Parallel (HSDP) setup. We further enable activation checkpointing at the unit of each DiT block. To improve training efficiency, we apply FlashAttention-3~\citep{shah2024flashattention} to the cross-attention layers and use FlexAttention~\citep{dong2024flex} to compile custom kernels for chunk-wise causal attention.

Training is conducted using BFloat16 precision and the AdamW optimizer~\citep{loshchilov2017decoupled}, with a learning rate of $1 \times 10^{-5}$, a cosine decay schedule, and a linear warm-up over the first 5\% of steps. Gradients are clipped to a maximum norm of 0.05. The model is trained for 5 epochs using 960 NVIDIA H200 Tensor Core GPUs.

\subsection{Stage 2: Joint Training}\label{si:stage2}

After the individual modules are prepared, we integrate them into a unified system and train them under a generative objective that couples latent prediction with reconstruction in the observation space.

A frequent concern over a generative scheme for world models is that the fine details of future observations are inherently unpredictable from past inputs. Directly reconstructing raw pixels often leads to unstable objectives dominated by high-frequency noise and stochastic variation. PAN treats this unpredictability not as evidence against generative modeling, but as a defining property of the physical world. In realistic world simulation, agents indeed routinely encounter novel viewpoints or unseen regions (e.g., turning around a corner or revealing an occluded object), yet coherence remains achievable if the underlying world dynamics are faithfully modeled. The goal of generative supervision in PAN is therefore not to reproduce the original world pixel-for-pixel, but to ensure that every generated observation remains semantically and physically consistent with the evolving latent world state.

This view is operationalized through the GLP framework, which separates structured world dynamics from unpredictable perceptual detail. Low-level stochastic variations are handled by the encoder--decoder pair $(h, g)$, while the predictive module $f$ focuses on modeling smooth and causally meaningful transitions in the latent space. The unified training objective is:
\begin{equation}
\mathcal{L}_{\text{GLP}}(h,f,g) = \mathbb{E}_{(o_t,a_t,o_{t+1})\sim\mathcal{D}} \big[\,\text{disc}\big(g \circ f(h(o_t),a_t),\, o_{t+1}\big)\,\big],
\end{equation}
where the discrepancy function $\text{disc}(\hat{o}_{t+1}, o_{t+1})$ is implemented as the flow-matching loss (\S\ref{si:flow-matching}). By grounding prediction in the observation space rather than matching latent embeddings, the training objective encourages each transition to correspond to a realizable sensory change and reduces susceptibility to the representation collapse that can occur under purely latent-space objectives~\citep{xing2025critiques}.

The backbone, built on Qwen2.5-VL-7B-Instruct, now produces latent world states that directly condition the decoder. This allows the decoder to learn to interpret and render the compact world state representations produced by the backbone, while the backbone itself adapts to produce states that are most effective for guiding high-fidelity, temporally coherent simulation. We freeze the vision-language model and train only the query embeddings and the video diffusion decoder. To align with the context window size of Qwen2.5-VL-7B-Instruct, we restrict the world model history to the most recent 10 rounds during training.

We reuse the HSDP strategy and activation checkpointing introduced in Stage~1 to manage memory consumption. Given the large sequence lengths in both the VLM and DiT, we apply sequence parallelism~\citep{li2021sequence} to shard hidden states across GPUs along the sequence length dimension. For self-attention modules, we adopt the Ulysses method~\citep{jacobs2023deepspeed}, which introduces all-to-all communication to change sharding across attention heads to calculate attention efficiently. To minimize communication overhead and maximize parallelism, we only use intra-node SP group with size 4, which is a common divisor of the 28 attention heads in Qwen2.5-VL-7B-Instruct and the 8 GPUs per compute node.

Similar to Stage~1, training is conducted with BFloat16 precision, the AdamW optimizer, a learning rate of $1 \times 10^{-5}$, cosine learning rate scheduling, and a 5\% warm-up. While the training schedule was set for 5 epochs, we perform early stopping after 1 epoch based on validation convergence.
During training, the model uses teacher-forced observed chunks to estimate latent histories and supervise next-chunk reconstruction. During inference, it performs closed-loop rollout by feeding back its own generated observations. The decoder condition $\hat{o}_t$ denotes the available visual context used by the video decoder, implemented through the sliding-window mechanism in Causal Swin-DPM (\S\ref{si:swdpm}) rather than through unrestricted access to all past frames.

\subsection{Inference}\label{si:inference}

During inference, PAN performs autoregressive multi-step simulation of world dynamics: given an initial observation image $o_1$, \modelname estimates the initial world state $\hat{s}_1 = h(o_1)$ and, based on proposed natural language actions $a_{1:T-1}$, predicts the future states $\hat{s}_{2:T}$ sequentially and reconstructs their corresponding observation videos $\hat{o}_{2:T}$.

To improve long-term consistency and generation fidelity, we augment the predicted states $\hat{s}_{2:t}$ with the encoder output of their corresponding reconstructed observations $\hat{o}_{2:t}$, resulting in enhanced future states $\hat{s}'_k = [\hat{s}_k, h(\hat{o}_k)]$ for $k \in [2, t]$. To better keep track of existing objects, agents, dynamics, and previous interactions, we concatenate the initial state with enhanced future states and all previous actions as $\tilde{s}_t = [\hat{s}_1, a_1, \hat{s}'_{2}, a_2, \dots, \hat{s}'_{t}]$ to serve as the input context for the world model backbone $f$, which predicts the next world state:
\begin{equation}
\hat{s}_{t+1} = f(\tilde{s}_t, a_t),
\end{equation}
Where the prediction is conditioned on the previously \textit{generated} observations
$\hat{o}_{2:t}$
rather than the ground truth, enabling open-loop simulation over extended horizons.
For the decoder, we include the previous observation $\hat{o}_t$ to better preserve perceptual continuity, and synthesize
the next observation from the predicted latent state as follows:
\begin{align}
    \hat{o}_{t+1} &= g(\hat{s}_{t+1}, \hat{o}_t). \label{eq:decoder_rollout}
\end{align}
This iterative process yields a coherent rollout of future world evolutions consistent across
both latent and perceptual domains. By conditioning every prediction on its own generated history,
PAN functions as a fully self-contained simulator capable of long-horizon imagination and reasoning.

\subsubsection{Causal Inference for Video Diffusion Decoder}

During the Causal Swin-DPM denoising process, the DiT output for the earlier chunk is computed at each step. Once the earlier chunk is fully denoised and dequeued, the cached output is discarded. When a new noisy chunk is enqueued, the model only needs to compute the output for the new chunk and reuse the cached output of the current chunk, reducing redundant computation.
During decoding, classifier-free guidance~\citep{ho2022classifier} is applied with a guidance scale of 4 to improve generation quality.

\subsubsection{Inference Accelerations}\label{si:accel}

\paragraph{Parallelization Strategy.}
To reduce the computational cost of inference, we apply sequence parallelism (SP) to split the video diffusion decoder's sequence across GPUs. The default SP group size is 8, and the chunk sequence is split unevenly across devices using a 5:3 ratio (first chunk : second chunk), reflecting the asymmetric attention pattern of the causal mask.

\paragraph{Quantized Attention.}
We further accelerate inference through quantized attention computation via SageAttention2++~\citep{zhang2025sageattention2++}. SageAttention2++ uses 8-bit compute precision with 4-bit key--value quantization to reduce memory bandwidth and computation costs. In our experiments, this quantization achieves a 30.3\% end-to-end speedup with negligible impact on generation quality.

\section{Training Data Pipeline}\label{si:data}

The dataset is drawn from videos in YouTube-8M~\citep{abu2016youtube} and YT-Temporal-180M \citep{zellers2021merlot}, two public accessible video datasets derived from public YouTube videos, spanning everyday human activities, human--object and human--environment interactions, natural and urban scenes, embodied navigation, and multi-agent scenarios. Our data strategy reflects a simple observation: any sequence that captures how the world evolves over time can serve as training material for next-state prediction. We therefore construct a dataset that represents the world through aligned video and text, ensuring that high-level semantic descriptions are paired with fine-grained visual dynamics in a way specifically tailored for world modeling.

Building a suitable training set for a general world model requires not only scale but also careful curation: the data must reflect rich temporal dynamics while being free of content that is uninformative or harmful for learning action-conditioned simulation.

\subsection{Video Segmentation}\label{si:segmentation}

Raw long-form videos are segmented into shorter clips using a dynamic shot-boundary detection strategy adapted from the Panda-70M pipeline~\citep{chen2024panda70m}. This three-stage approach detects both hard cuts (abrupt scene transitions) and soft transitions (gradual fades, dissolves), producing clips that correspond to coherent visual episodes with consistent scene structure and temporal continuity.

\subsection{Data Filtering Pipeline}\label{si:filtering}

After segmentation, we apply a multi-stage filtering pipeline to remove clips unsuitable for world-model training. The pipeline consists of three complementary stages: rule-based motion and quality filters, pretrained detector-based filters, and a custom VLM-based filter.

Supplementary Table~\ref{tab:filtering} summarizes the filtering statistics at each stage.

\begin{table}[t]
\centering
\begin{tabular}{llc}
\toprule
\textbf{Filtering Stage} & \textbf{Metric} & \textbf{Filtered Ratio} \\
\midrule
\multirow{7}{*}{Extremely Static or Overly Dynamic} & Low luminance change & 5.4\% \\
& Low edge difference  & 2.3\% \\
& Excessive luminance change & 5.4\% \\
& Unstable Dynamics & 20.4\% \\
& Scene Cut & 30.3\% \\
& Low OFM & 12.1\% \\
& Overall & 47.0\% \\
\midrule
\multirow{6}{*}{Trivial Motion or Pure Color} & Translation & 26.6\% \\
& Zoom In & 6.9\% \\
& Zoom Out & 4.3\% \\
& Color (first frame) & 0.03\% \\
& Color (last frame) & 0.05\% \\
& Overall & 30.8\% \\
\midrule
Low Aesthetic Quality & Overall & 27.0\% \\
\midrule
\multirow{3}{*}{Obstructive Text} & Average across frames & 1.6\% \\
& Delta across frames & 1.8\% \\
& Overall & 3.1\% \\
\midrule
\multirow{6}{*}{VLM-based Filters} & Lecture type & 26.4\% \\
& Text dominated & 12.3\% \\
& Noisy screenshot & 1.4\% \\
& Low quality & 0.6\% \\
& Edited & 6.0\% \\
& Scene cut & 9.7\% \\
& Overall & 37.9\% \\
\midrule
Overall Filtered & & 83.9\% \\
\bottomrule
\end{tabular}

\caption{Percentage of filtered clips at each filtering stage. The percentage at each stage is computed over clips that passed all previous stages. A single clip may be flagged by multiple metrics within the same stage.}
\label{tab:filtering}

\end{table}

\subsubsection{Rule-Based Filters}

The first stage applies motion-based and statistical heuristics to remove clips with uninformative temporal dynamics.

\paragraph{Extremely static and dynamic clips.} We compute three per-frame descriptors: Optical Flow Magnitude (OFM), edge difference, and luminance difference, to characterize the temporal dynamics of each clip. Clips are classified into six rejection categories based on these descriptors:
\begin{enumerate}[leftmargin=*,label=(\roman*)]
\item \textit{extremely static} (negligible OFM and edge/luminance change);
\item \textit{extremely dynamic} (excessive OFM and/or edge/luminance change);
\item \textit{static-then-dynamic} (motion concentrated at the end);
\item \textit{unstable dynamics} ($\mu_{\Delta L} < \sigma_{\Delta L}$ and $\mu_{\Delta E} < \sigma_{\Delta E}$, capturing noisy motion);
\item \textit{scene cut} (any frame transition with $\Delta L$ or $\Delta E$ exceeding a high threshold); and
\item \textit{low OFM} (average OFM below a set threshold).
\end{enumerate}
For shot boundary detection within this stage, we use the \texttt{scenedetect} library.

\paragraph{Trivial motion and pure color.} We further remove clips dominated by trivial or uninformative motion patterns, such as uniform camera translation or zooming. We take the middle frame of each clip as a reference and track up to 1,000 Shi--Tomasi keypoints both forward and backward in time using a pyramidal Lucas--Kanade tracker, then aggregate the resulting displacement vectors into translation and zoom scores. Clips exhibiting highly regular motion according to these measures are filtered out. Because zoom-in motion introduces less new visual information than zoom-out, we apply a stricter threshold to zoom-in than to zoom-out. In addition, videos that begin or end with nearly uniform color frames, often caused by fade-in or fade-out transitions or blank screens, are excluded.
Specifically, we take the middle frame of each clip as a reference and track up to 1{,}000 Shi--Tomasi keypoints both forward and backward in time using a pyramidal Lucas--Kanade tracker, then aggregate the resulting displacement vectors into the translation and zoom scores.
Because zoom-in motion introduces less new visual information than zoom-out, we apply a stricter threshold to zoom-in than to zoom-out.

\subsubsection{Pretrained Detectors}\label{si:detectors}

Next, we employ existing pretrained models to identify quality issues that are difficult to capture with simple heuristics.

\paragraph{Low aesthetic quality.} Training on visually appealing and well-composed videos improves the quality and stability of generation. The aesthetic scorer~\citep{schuhmann2022improved_aesthetic_predictor} consists of a pretrained CLIP ViT-L/14 model followed by an MLP regressor trained on the AVA dataset~\citep{murray2012ava}. We uniformly sample several frames per clip and average the frame-level predictions; clips with scores below a defined threshold are excluded.
Concretely, the aesthetic scorer~\citep{schuhmann2022improved_aesthetic_predictor} consists of a pretrained CLIP ViT-L/14 model followed by an MLP regressor trained on the AVA dataset~\citep{murray2012ava}.

\paragraph{Obstructive text.} Videos that contain large or persistent text regions, such as subtitles, banners, or watermarks, often distract the model and can lead to undesired text artifacts. We adopt MixNet~\citep{zeng2023mixnetaccuratedetectionchallenging}, a robust scene-text detector: for each video we sample the start, middle, and end frames, detect text regions, and compute the total text-covered area normalized by the image area, excluding any clip whose text-area ratio exceeds a predefined threshold.

\subsubsection{Custom VLM-Based Filter}\label{si:vlm-filter}

Finally, we train a custom vision--language model (VLM) filter to identify categories of low-quality content that earlier stages cannot reliably capture. The VLM is designed to detect and exclude:
\begin{enumerate}[leftmargin=*]
\item \textbf{Lecture-type videos}: clips in which a person speaks directly to the camera without performing any meaningful actions.
\item \textbf{Text-dominated videos}: videos where large text banners or overlays occupy a significant portion of the visual content.
\item \textbf{Screen recordings or noisy screenshots}: clips showing desktop or mobile interfaces that lack real-world dynamics.
\item \textbf{Low-quality content}: videos with severe blurriness, compression artifacts, or other forms of visual degradation.
\item \textbf{Heavily edited clips}: videos that contain transitions, animations, or special effects not useful for modeling natural world dynamics.
\item \textbf{Residual scene cuts}: clips with abrupt transitions or compositional changes that remain after the initial segmentation step.
\end{enumerate}

We randomly sampled 11,173 clips that had passed the rule-based filters and pretrained detectors and asked human annotators to label them according to the categories above, then fine-tuned a Qwen2.5-VL-7B model~\citep{bai2025qwen2} on this set using a 9:1 train/validation split. For each property, the filtering score is the softmax probability of the ``yes'' token at the corresponding output position, and its threshold is chosen to maximize the $F_1$ score on the validation set; a clip is discarded if any property exceeds its threshold.

\subsection{Dense Video Captioning}\label{si:captioning}

The collected videos either lack captions or include only brief descriptions that fail to capture the rich visual content within each clip. We identify two key requirements for captions: (1) they must be factually rich and detailed rather than high-level or generic, following findings from DALL-E~3~\citep{betker2023improving}; and (2) they must focus on temporal dynamics, including motion, events, changes in the environment, and the emergence of new objects, rather than static background elements. The static content is typically available in the initial frame or previous video state, while the evolving dynamics correspond to the action inputs of the world model and drive state transitions. To meet these requirements, we re-caption the video data using a vision--language model specifically prompted to generate dense, temporally grounded descriptions. This yields 8 million video--text pairs corresponding to approximately 4,500 hours of video.

\section{Evaluation Details}\label{si:eval}

\subsection{Benchmark Specifications}
\label{si:eval-benchmark}

We adopt the three-dimensional evaluation framework from WR-Arena~\citep{gao2026wrarena}. Full protocol details, metric definitions, dataset specifications, and evaluation code are available in~\citet{gao2026wrarena}. We summarize each dimension below.

\paragraph{Action Simulation Fidelity.} Given an initial image observation, GPT-4o~\citep{openai2024gpt4technicalreport} proposes several feasible, non-contradictory action sequences. The world model simulates corresponding rollouts, which are scored by a VLM-based judge along the criteria of action faithfulness and precision, following the protocol of~\citet{worldmodelbench}. Two settings are evaluated: \textit{Agent Simulation}, in which the model drives a controllable entity according to specified behaviors while maintaining background stability across multiple counterfactual action sequences from the same initial state; and \textit{Environment Simulation}, in which the model performs scene-level interventions such as adding, removing, or moving objects, or changing weather and lighting conditions.

\paragraph{Long-Horizon Forecast.} Starting from an initial observation and a multi-step action script, the protocol measures two complementary aspects: \textit{Transition Smoothness}, which quantifies local motion continuity at chunk boundaries using dense optical flow velocity and acceleration, rewarding smooth motion evolution and penalizing abrupt transitions; and \textit{Simulation Consistency}, which tracks content alignment and style consistency over extended rollouts using metrics adapted from WorldScore~\citep{duan2025worldscore}, with progressive penalties that emphasize late-stage prediction quality.

\paragraph{Simulative Reasoning and Planning.} This dimension evaluates whether a world model can serve as an internal simulator that enables an agent to reason about actions and plan toward a goal. Three settings are evaluated: \textit{Step-Wise Simulation} measures atomic causal prediction in robotic manipulation scenarios from Agibot~\citep{bu2025agibot}, following WM-ABench~\citep{gao2025visionlanguagemodelsinternalworld}; \textit{Open-Ended Simulation and Planning} evaluates goal-directed manipulation on 15 diverse scenarios drawn from Agibot, where a VLM agent, OpenAI-o3~\citep{OpenAI2025o3o4minisystemcard}, proposes candidate actions, the world model simulates their consequences, and the agent selects the action whose predicted observation most closely approaches the goal; \textit{Structured Simulation and Planning} applies the same agent--world-model loop to 47 test cases in tabletop environments from the Language Table dataset~\citep{lynch2023interactive}.

\subsection{Baseline Descriptions}
\label{si:eval-baseline}

\textbf{WAN 2.1-I2V-14B} is a large diffusion-transformer model for high-fidelity image-to-video generation. It focuses on photorealistic rendering and consistent scene dynamics.

\textbf{WAN 2.2-I2V-14B} is an iterative successor to WAN 2.1 that improves visual fidelity and temporal stability while expanding controllable prompting capabilities. It adopts a mixture-of-experts architecture.

\textbf{Cosmos1-14B} is NVIDIA's first-generation World Foundation Model that learns physical dynamics and object affordances from large-scale video corpora. It predicts future states for action-conditioned rollouts.

\textbf{Cosmos2-14B} scales capacity and controllability over Cosmos1, improving long-horizon coherence and physics fidelity. It is designed to be post-trained for domain applications such as camera control, robotic manipulation, and autonomous driving.

\textbf{V-JEPA 2} extends Meta's Joint Embedding Predictive Architecture to video. Rather than predicting pixels, it predicts future latent representations, functioning as a compact latent world model rather than a video generator.

\textbf{KLING} is a production-grade text- and image-to-video model developed by Kuaishou. It produces high-quality 1080p outputs with controllable camera motion and temporal coherence.

\textbf{MiniMax-Hailuo} is a commercial video-generation API designed for high-resolution, minute-scale video synthesis. It supports common frame rates and user-prompted creative controls.

\textbf{Gen-3} is Runway's third-generation model emphasizing reliability, character consistency, and fine-grained spatiotemporal control.

\section{Related Work}\label{si:related}

\paragraph{World models and latent dynamics.}
The concept of a learned world model, namely a system that predicts future states given actions, has a long history in cognitive science and AI~\citep{tolman1948cognitive,briscoe2011mental,battaglia2013simulation}.
Modern computational world models trace to~\citet{ha2018world}, who combined a variational autoencoder with an RNN-based dynamics model for policy learning in latent space.
The Dreamer line of work~\citep{hafner2019dream,hafner2020mastering,hafner2023mastering} extended this paradigm to achieve competitive performance on continuous control tasks through latent imagination and actor-critic learning, demonstrating that learned dynamics models can support effective planning.
PlaNet~\citep{hafner2019learning} showed that planning entirely in latent space, without a learned policy, can solve visual control tasks.
IRIS~\citep{micheli2022transformers} and DIAMOND~\citep{alonso2024diffusion} applied discrete tokenization and diffusion-based generation, respectively, to Atari-scale world modeling.
MuZero~\citep{schrittwieser2020mastering} demonstrated that a learned dynamics model can support planning in board games and Atari without access to environment rules.
These works establish the encoder--dynamics--decoder decomposition that \modelname adopts and develops through the GLP architecture~\citep{xing2025critiques}.
\modelname's specific contribution relative to this lineage is the combination of an LLM backbone with a video diffusion decoder for open-domain, language-conditioned simulation, rather than domain-specific control.

\paragraph{Interactive and domain-specific world models.}
A growing body of work targets interactive simulation in specific domains.
In gaming, Genie~\citep{bruce2024genie} learns playable environments from unlabeled video, while MineWorld~\citep{guo2025mineworld}, Oasis~\citep{decart2024oasis}, and Genie 2/3~\citep{parker2024genie,genie3} generate interactive environments from images or text.
In autonomous driving, GAIA-1~\citep{hu2023gaia}, DriveDreamer~\citep{wang2023drivedreamer,zhao2024drivedreamer}, and the Cosmos series~\citep{cosmos,nvidia-cosmos-predict2,nvidia2026cosmos3} learn action-conditioned forecasting from driving data.
In robotics, UniSim~\citep{yang2024learning} trains a general-purpose simulator from diverse real-world data, and several models~\citep{zhou2024robodreamer,du2023learning} predict future visual states for robotic manipulation.
3D world models such as World Labs~\citep{worldlabs2024generating,worldlabs2025marble} focus on scene generation and egocentric navigation, but typically emphasize static spatial structure over dynamic physics and action-conditioned rollout.
These domain-specific models achieve strong performance within their target environments.
\modelname aims for a complementary capability: broad-domain simulation under language-level actions, trading task-specific precision for domain generality.

\paragraph{Joint embedding predictive models.}
The JEPA framework~\citep{lecun2022path} and its instantiations: V-JEPA~\citep{bardes2024revisiting}, V-JEPA 2~\citep{assran2025vjepa2selfsupervisedvideo}, and DINO-WM~\citep{zhoudino}, represent an influential alternative that predicts future latent states without reconstructing observations.
This design avoids modeling fine-grained visual details and can yield efficient representations for downstream tasks.
As analyzed in~\citet{xing2025critiques}, however, purely latent-space objectives provide a looser surrogate for observation-level consistency, and may not retain all information needed for visual simulation.
\modelname takes a complementary direction by grounding predicted transitions through generative reconstruction, which is especially important for action-conditioned video rollout where the model must render the observable consequences of candidate actions.

\paragraph{Video generation models.}
Recent video generation models have achieved striking visual realism \citep{videoworldsimulators2024,veo2024,kong2024hunyuanvideo,wan2025wan,yang2024cogvideox}.
Diffusion Transformer (DiT) architectures~\citep{peebles2023scalable} have enabled scaling to longer, higher-quality outputs.
While some of these systems have been described as world simulators, they typically produce complete videos from fixed prompts without the explicit action conditioning, sequential state evolution, or closed-loop rollout that characterize world models.
\modelname builds on the visual fidelity of video generation models, specifically adapting Wan2.1-T2V-14B as its decoder, while adding the structured dynamics backbone and action-conditioned rollout needed for world simulation.

\paragraph{Hybrid LLM--diffusion architectures.}
The architectural pattern of using LLM predictions to condition diffusion decoders has been explored in multimodal generation.
EMU~\citep{sun2024generative} and GILL~\citep{koh2023generating} use LLM-generated embeddings to condition image generation for interleaved text-image outputs.
MM-Interleaved~\citep{tian2024mm} synchronizes multimodal features across LLM and diffusion components.
MarDini~\citep{liu2024mardini} and HART~\citep{tang2025hart} combine autoregressive planning with lightweight diffusion heads for video and image generation, respectively.
These works demonstrate the viability of the LLM--diffusion coupling for content generation.
\modelname applies this architectural pattern to a different purpose: world simulation rather than content generation, with the LLM backbone modeling temporal dynamics conditioned on actions rather than generating diverse media from prompts.

\paragraph{World model evaluation.}
Systematic benchmarking of world models is an active and evolving area.
WM-ABench~\citep{gao2025visionlanguagemodelsinternalworld} evaluates causal prediction in robotic manipulation; WorldScore~\citep{duan2025worldscore} measures visual consistency over extended rollouts; Cosmos Bench~\citep{nvidia_cosmos_bench} assesses physical understanding in driving scenarios.
Standard reinforcement-learning benchmarks such as Atari~\citep{bellemare2013arcade} and the DeepMind Control Suite~\citep{tassa2018deepmind} remain central to evaluating world models in closed-loop control settings with low-dimensional action spaces.
WR-Arena~\citep{gao2026wrarena} extends this landscape by measuring action simulation fidelity, long-horizon forecast, and simulative reasoning and planning under open-domain, language-conditioned rollout, targeting the evaluation regime relevant to models such as \modelname.

\section{Additional Analysis}
\label{si:additional-analysis}

\subsection{Error Accumulation Over Long-Horizon Rollouts}
\label{si:error-accumulation}

\begin{figure}[t]
\centering
\includegraphics[width=0.8 \linewidth]{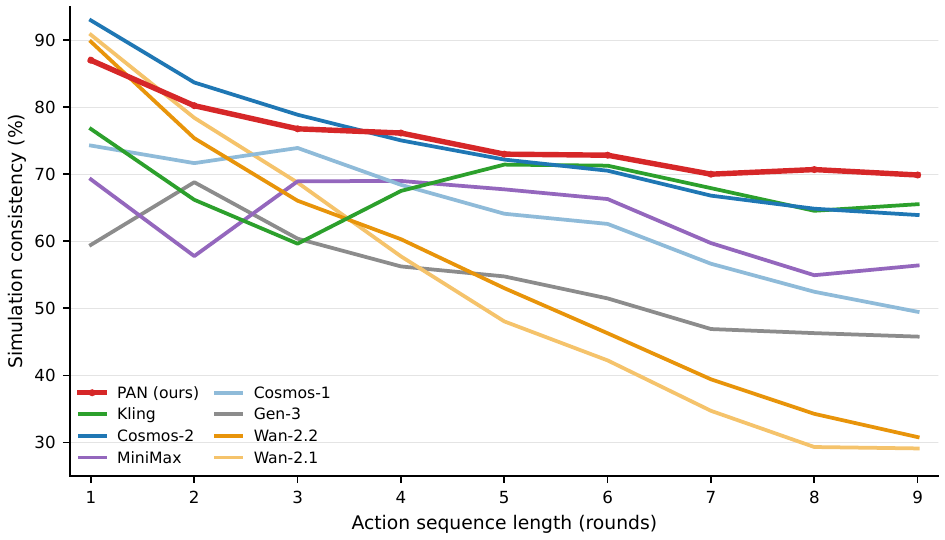}
\caption{\textbf{Error accumulation over long-horizon rollouts.}
Simulation consistency (\%) measured at each round as action sequences extend
from 1 to 9 steps. Each curve corresponds to one model; higher is better.
PAN degrades gracefully and attains the highest consistency from round 5
onwards, whereas standalone video generators such as Wan-2.1 and Wan-2.2,
despite strong single-step scores, deteriorate rapidly as the horizon grows.}
\label{fig:si-error-accumulation}
\end{figure}

A central requirement for a world model is that its predictions remain
faithful as simulations extend over many action steps, since errors made at
each step compound through the autoregressive rollout. To characterize this
behaviour, we measure simulation consistency at every round of interaction as
action sequences extend from 1 to 9 steps, for PAN and for the baseline
models described in \S\ref{si:eval-baseline}. Results are shown in
Figure~\ref{fig:si-error-accumulation}.

Three observations emerge. First, all models lose consistency as the horizon
grows, reflecting the compounding of per-step prediction error, but the rate
of degradation varies widely. PAN declines gradually from 87.0 at round 1 to
69.9 at round 9, with the curve flattening over the final rounds, and it
attains the highest consistency of all models from round 5 onwards.
Second, strong single-step generation does not imply long-horizon stability.
Cosmos-2 achieves the best first-round score (93.0) but degrades more steeply
than PAN, falling below it by round 5 and ending at 63.9. The contrast is
sharpest for the standalone video generators Wan-2.1 and Wan-2.2, which open
at 90.8 and 89.8, respectively, yet collapse to 29.1 and 30.8 by round 9,
losing roughly two thirds of their initial consistency.
Third, the comparison with Wan-2.1 is particularly informative because PAN's
video diffusion decoder is initialized from this model
(\S\ref{si:arch}).

The two systems therefore share the same
underlying generative capacity, yet exhibit opposite long-horizon behaviour.
This indicates that PAN's stability under extended rollouts is not inherited
from the decoder, but arises from the surrounding framework: the Causal
Swin-DPM mechanism in the decoder that improves long-horizon temporal
continuity, and the autoregressive latent dynamics backbone that maintains a
persistent world state across steps
(\S\ref{si:arch}).

To isolate the contributions of PAN's key components, we conduct a
progressive ablation at 14B scale across three configurations that mirror
the stages of PAN's training procedure:
(i)~Wan-2.1, the base image-to-video model with
single-frame conditioning; (ii)~PAN\textsubscript{no-vlm}, the video
diffusion model trained on PAN's data pipeline under Causal Swin-DPM
supervision, without the VLM backbone;
and (iii)~the full PAN system
(\S\ref{si:swdpm-ablation}--\ref{si:vlm-ablation}).

\subsection{Ablation of the Causal Swin-DPM Denoising Mechanism}
\label{si:swdpm-ablation}

To ablate Stage~1 training and the Causal Swin-DPM denoising mechanism
it introduces, we compare PAN\textsubscript{no-vlm} (i.e., the model trained in
Stage~1 under Causal Swin-DPM supervision) against Wan-2.1, the pretrained
base model before Stage~1 adaptation. Both systems share the same underlying
model structure; the key difference is that Wan-2.1
relies on single last-frame conditioning for sequential generation, whereas
Causal Swin-DPM trains the model to denoise each chunk conditioned on
multiple partially noised frames from the preceding chunk. Results are
reported in Table~\ref{tab:si-vlm-ablation}.

Transition
Smoothness improves from 11.0 to 44.6, indicating smooth visual transitions
at chunk boundaries. Simulation Consistency improves from 36.2 to 60.9,
reflecting suppressed accumulation of artifacts and motion drift over long
rollouts. These improvements are driven by the Causal Swin-DPM denoising
mechanism: the model learns to attend to partially denoised
context from the previous chunk instead of a single fully denoised final
frame. This partially noised context carries persistent scene structure,
object presence, and motion direction rather than incidental pixel-level
details. It improves transition smoothness at chunk boundaries, and dampens
error accumulation across rollout steps.
The conditioning also carries a richer representation of the current
scene state than a single final frame; combined with the
world-modeling-oriented data pipeline that aligns actions with this
evolving state, it improves Action Simulation Fidelity
(Overall: 45.3 to 56.8).

These results confirm that Causal Swin-DPM generative supervision, in combination with PAN's training process,
contributes substantially to PAN's transition continuity and long-horizon
stability.

\subsection{Ablation of the VLM-based Latent Dynamics Backbone}
\label{si:vlm-ablation}

To isolate the contribution of the latent dynamics backbone, we compare the
full PAN model against PAN\textsubscript{no-vlm}, the ablated variant from
\S\ref{si:swdpm-ablation}

in which the VLM-based backbone is removed
entirely and the video diffusion decoder is conditioned directly on the
observation history and the language-specified action. This variant retains
PAN's generative capacity, Causal Swin-DPM supervision, and training data,
but lacks an explicit latent world-state representation.
Table~\ref{tab:si-vlm-ablation} reports results on both evaluation
dimensions.

\begin{table}[t]
\centering
\caption{\textbf{Ablation of the Causal Swin-DPM denoising mechanism and VLM-based latent dynamics backbone.}
Wan-2.1 is the base image-to-video pretrained base model of PAN decoder. PAN\textsubscript{no-vlm} removes the backbone and conditions the video
diffusion decoder directly on observations and actions. Best results in bold.}
\label{tab:si-vlm-ablation}
\medskip
\begin{tabular}{lccccc}
\toprule
& \multicolumn{3}{c}{Action Simulation Fidelity}
& \multicolumn{2}{c}{Long-Horizon Forecast} \\
\cmidrule(lr){2-4}\cmidrule(lr){5-6}
Model & Environment & Agent & Overall
& \makecell{Transition\\Smoothness} & \makecell{Simulation\\Consistency} \\
\midrule
Wan-2.1 & 36.0 & 53.7 & 45.3 & 11.0 & 36.2 \\

PAN\textsubscript{no-vlm}   & \textbf{47.7} & 66.0 & 56.8 & 44.6 & 60.9 \\
PAN                         & 47.1 & \textbf{70.3} & \textbf{58.7}
                            & \textbf{53.6} & \textbf{64.1} \\
\bottomrule
\end{tabular}
\end{table}

The pattern of gains is informative about what the backbone contributes.
Environment simulation is essentially unchanged (47.7 versus 47.0),
suggesting that rendering the visual dynamics of the environment itself is
carried largely by the diffusion decoder and does not depend on the backbone.
In contrast, agent simulation improves by 4.3 points with the backbone in
place. Simulating the behaviour of agents in the scene requires interpreting
the language-specified action, relating it to the entities present, and
inferring plausible responses, all of which are capabilities supported by the world knowledge
and language grounding acquired during VLM pretraining, and which the
decoder-only variant lacks.
The largest improvement appears in transition smoothness, which rises by 9.0
points (44.6 to 53.6). Generating smooth transitions across chunk boundaries
requires that successive predictions be produced from a coherent, persistent
representation of the world state rather than re-derived from raw pixels at
each step. The latent backbone provides exactly this continuity: each chunk
is decoded conditioned on the evolving latent state, so the model carries
forward information about scene layout, object identity, and ongoing motion
that a purely observation-conditioned generator must repeatedly re-infer.
Simulation consistency also improves, by a smaller margin (60.9 to 64.1),
consistent with the long-horizon analysis above: the ablated variant, trained
under the same framework, remains far more stable than standalone video
generators, while the backbone contributes an additional gain concentrated in
the coherence of transitions and of agent behaviour.

Taken together, the two analyses attribute PAN's long-horizon stability to
complementary components: the training framework and chunk-wise generation
mechanism prevent the rapid collapse observed in standalone video generators,
while the VLM-based latent backbone is chiefly responsible for coherent
transitions and faithful agent dynamics.

\subsection{Training Dynamics and Loss Trajectory}
\label{si:training-dynamics}

\begin{figure}[t]
\centering
\includegraphics[width=0.8\linewidth]{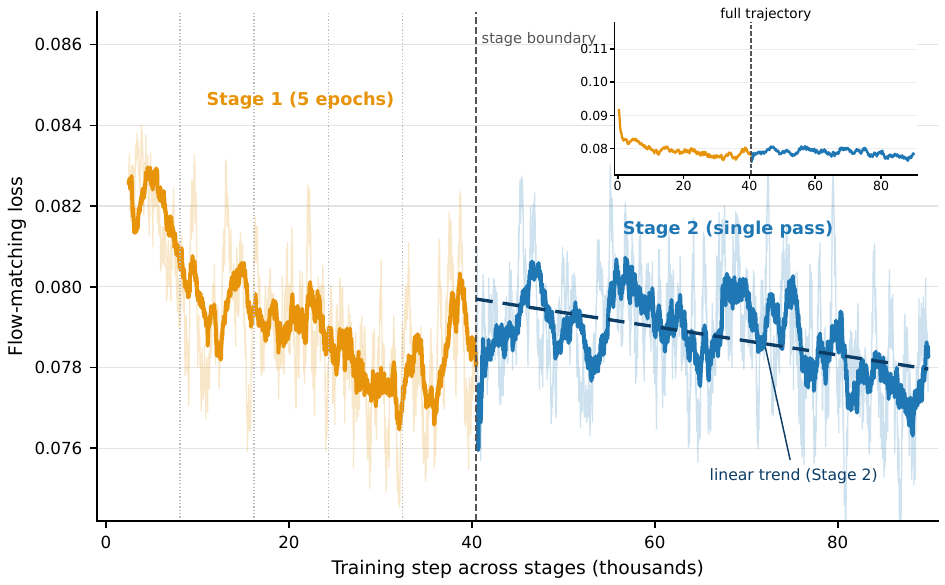}
\caption{\textbf{Training loss trajectory across both stages.}
Flow-matching loss of the full system over Stage 1 (orange, five epochs;
dotted lines mark epoch boundaries) and Stage 2 (blue, single pass; the
dashed vertical line marks the stage boundary), shown excluding the initial
adaptation phase; the inset shows the full trajectory. Light traces show a
400-step rolling mean of the raw per-step loss, whose variance stems from
the sampling of diffusion timesteps; heavy lines show a 1{,}500-step rolling
mean. The dashed dark line is a linear fit to the raw Stage 2 loss over the
entire pass. Step counts are not comparable across stages owing to different
batch sizes.}
\label{fig:si-loss-trajectory}
\end{figure}

Figure~\ref{fig:si-loss-trajectory} shows the flow-matching training loss of
the full PAN system over the course of both training stages
(\S\ref{si:training}):

Stage 1, in which the video diffusion
decoder is adapted over five epochs of the training corpus, and Stage 2, in
which the backbone and decoder are trained jointly under the GLP objective
for a single pass over data drawn from the same curated corpus
(\S\ref{si:data}).

The two stages use different batch
configurations, so step counts are not directly comparable across stages;
the horizontal axis should be read as training progress within each stage.

The trajectory exhibits three properties relevant to interpreting PAN's
results. First, training is stable throughout. Stage 1 loss falls rapidly
during initial adaptation (from 0.116 to below 0.085 early in the first
epoch) and drifts gradually lower over the remaining epochs, with mild
fluctuations but no discontinuities at epoch boundaries; the absence of
abrupt drops when the data repeats indicates that the decoder is fitting the
underlying dynamics
of the corpus rather than memorizing individual sequences.
Second, the transition to joint training is seamless: Stage 2 loss resumes
at the level where Stage 1 ends and remains stable as the backbone enters
the training loop, indicating that the GLP objective can be introduced on
top of the adapted decoder without destabilizing optimization.
Third, and most relevant to the question of scale, the loss has not
saturated by the end of training. A linear fit to the raw per-step loss over
the entire single pass of Stage 2 yields a steadily negative trend,
corresponding to a reduction of approximately 2\% in loss from the beginning
to the end of the pass (Figure~\ref{fig:si-loss-trajectory}). The system is
therefore still improving, at a gradual but consistent rate, at the point
where training concludes. While a full characterization of PAN's scaling
behaviour would require training runs at multiple data and compute budgets,
which we leave to future work, the persistent downward trend indicates that
the reported results do not reflect a saturated model, and that additional
data and compute offer a direct avenue for further improvement.

\section{Limitations and Future Work}
\label{sec:limitations}
\paragraph{Language-level action interface.}
\modelname's current action interface is based on natural-language instructions at the video-chunk level.
This makes the model suitable for open-domain, high-level action-conditioned rollouts and counterfactual visual imagination (e.g., simulating the consequences of ``pick up the cup'' or ``turn left at the intersection''), but less suitable for tasks requiring precise low-level control, such as continuous torque commands, exact navigation waypoints, or closed-loop manipulation policies.
In practice, complex physical interactions that require specifying force, direction, and timing simultaneously can be inefficient to express in natural language, and \modelname's action fidelity on such tasks is correspondingly lower.
Future versions should support richer action representations, including continuous controls, robot proprioception, and structured environment feedback.

\paragraph{Long-horizon drift and failure modes.}
While Causal Swin-DPM improves temporal continuity across generated chunks, long-horizon simulation remains an open challenge.
\modelname can still exhibit semantic drift, object inconsistency, accumulation of small visual artifacts, and reduced action controllability as rollouts extend or move away from the training distribution.
These failure modes are especially visible in scenes involving small objects, precise spatial relations, rare physical interactions, crowded multi-agent environments, or actions whose effects are delayed over many steps.
The long-horizon results reported in this work should therefore be interpreted as evidence of improved rollout stability under the evaluated settings, rather than as a complete solution to indefinite-horizon world simulation.

\paragraph{Inherited limitations from the video generation backbone.}
\modelname builds on Wan~2.1, an open-source video generation model originally designed for high-quality content creation rather than physically grounded simulation.
Although adapting such a model provides a strong visual prior and enables efficient scaling, it also inherits a stronger emphasis on perceptual realism than on causal correctness, physical consistency, and controllable state evolution.
For example, the model may produce visually plausible but physically incorrect object interactions (e.g., objects passing through surfaces) or may hallucinate scene elements not implied by the action instruction.
A natural next step is to train a world model from scratch with objectives, architectures, and data specifically designed for simulation, interaction, and long-horizon predictive consistency.

\clearpage
\bibliographystyle{plainnat}
\bibliography{references}

@article{ye2026dreamzero,
  title        = {World Action Models are Zero-shot Policies},
  author       = {Ye, Seonghyeon and Ge, Yunhao and Zheng, Kaiyuan and Gao, Shenyuan and Yu, Sihyun and Kurian, George and Indupuru, Suneel and Tan, You Liang and Zhu, Chuning and Xiang, Jiannan and Malik, Ayaan and Lee, Kyungmin and Liang, William and Ranawaka, Nadun and Gu, Jiasheng and Xu, Yinzhen and Wang, Guanzhi and Hu, Fengyuan and Narayan, Avnish and Bjorck, Johan and Wang, Jing and Kim, Gwanghyun and Niu, Dantong and Zheng, Ruijie and Xie, Yuqi and Wu, Jimmy and Wang, Qi and Julian, Ryan and Xu, Danfei and Du, Yilun and Chebotar, Yevgen and Reed, Scott and Kautz, Jan and Zhu, Yuke and Fan, Linxi and Jang, Joel},
  journal      = {arXiv preprint arXiv:2602.15922},
  year         = {2026},
}

@article{nvidia2026cosmos3,
  title = {Cosmos 3: Omnimodal World Models for Physical AI},
  author = {{NVIDIA}},
  journal = {arXiv preprint arXiv:2606.02800},
  year = {2026},
  url = {https://research.nvidia.com/labs/cosmos-lab/cosmos3/technical-report.pdf}
}

@misc{li2026functional_taxonomy_world_models,
  author       = {Li, Fei-Fei},
  title        = {A Functional Taxonomy of World Models},
  year         = {2026},
  month        = jun,
  day          = {3},
  howpublished = {X post},
  url          = {https://x.com/drfeifei/status/2062247238143996275},
  note         = {Accessed: 2026-06-05}
}

@article{deng2025general,
  title={General Agentic Planning Through Simulative Reasoning with World Models},
  author={Deng, Mingkai and Hou, Jinyu and Hu, Zhiting and Xing, Eric},
  journal={arXiv preprint arXiv:2507.23773},
  year={2025}
}

@article{zhang2025fast,
  title={Fast video generation with sliding tile attention},
  author={Zhang, Peiyuan and Chen, Yongqi and Su, Runlong and Ding, Hangliang and Stoica, Ion and Liu, Zhengzhong and Zhang, Hao},
  journal={arXiv preprint arXiv:2502.04507},
  year={2025}
}

@article{zhang2026faster,
  title={Faster video diffusion with trainable sparse attention},
  author={Zhang, Peiyuan and Chen, Yongqi and Huang, Haofeng and Lin, Will and Liu, Zhengzhong and Stoica, Ion and Xing, Eric and Zhang, Hao},
  journal={Advances in Neural Information Processing Systems},
  volume={38},
  pages={152509--152534},
  year={2026}
}

@article{xing2026critique,
  title={Critique of Agent Model},
  author={Xing, Eric and Deng, Mingkai and Hou, Jinyu},
  journal={arXiv preprint arXiv:2606.23991},
  year={2026}
}

@article{micheli2022transformers,
  title={Transformers are sample-efficient world models},
  author={Micheli, Vincent and Alonso, Eloi and Fleuret, Fran{\c{c}}ois},
  journal={arXiv preprint arXiv:2209.00588},
  year={2022}
}

@article{legg2007universal,
  title={Universal intelligence: A definition of machine intelligence},
  author={Legg, Shane and Hutter, Marcus},
  journal={Minds and machines},
  volume={17},
  number={4},
  pages={391--444},
  year={2007},
  publisher={Springer}
}

@article{tassa2018deepmind,
  title={Deepmind control suite},
  author={Tassa, Yuval and Doron, Yotam and Muldal, Alistair and Erez, Tom and Li, Yazhe and Casas, Diego de Las and Budden, David and Abdolmaleki, Abbas and Merel, Josh and Lefrancq, Andrew and others},
  journal={arXiv preprint arXiv:1801.00690},
  year={2018}
}

@article{bellemare2013arcade,
  title={The arcade learning environment: An evaluation platform for general agents},
  author={Bellemare, Marc G and Naddaf, Yavar and Veness, Joel and Bowling, Michael},
  journal={Journal of artificial intelligence research},
  volume={47},
  pages={253--279},
  year={2013}
}

@article{gao2026wrarena,
  title={World Reasoning Arena},
  author={Gao, Qiyue and Zhou, Kun and Xiang, Jiannan and Liu, Zihan and Yang, Dequan and Chen, Junrong and Ahmad, Arif and Zeng, Cong and Bannur, Ganesh and Huang, Xinqi and Liu, Zheqi and Gu, Yi and Yang, Yichi and Liu, Guangyi and Hu, Zhiting and Liu, Zhengzhong and Xing, Eric},
  journal={arXiv preprint arXiv:2603.25887},
  year={2026}
}

@inproceedings{tang2025hart,
  title={Hart: Efficient visual generation with hybrid autoregressive transformer},
  author={Tang, Haotian and Wu, Yecheng and Yang, Shang and Xie, Enze and Chen, Junsong and Chen, Junyu and Zhang, Zhuoyang and Cai, Han and Lu, Yao and Han, Song},
  booktitle={International Conference on Learning Representations},
  volume={2025},
  pages={68413--68432},
  year={2025}
}

@article{liu2024mardini,
  title={Mardini: Masked autoregressive diffusion for video generation at scale},
  author={Liu, Haozhe and Liu, Shikun and Zhou, Zijian and Xu, Mengmeng and Xie, Yanping and Han, Xiao and P{\'e}rez, Juan C and Liu, Ding and Kahatapitiya, Kumara and Jia, Menglin and others},
  journal={arXiv preprint arXiv:2410.20280},
  year={2024}
}

@article{tian2024mm,
  title={Mm-interleaved: Interleaved image-text generative modeling via multi-modal feature synchronizer},
  author={Tian, Changyao and Zhu, Xizhou and Xiong, Yuwen and Wang, Weiyun and Chen, Zhe and Wang, Wenhai and Chen, Yuntao and Lu, Lewei and Lu, Tong and Zhou, Jie and others},
  journal={arXiv preprint arXiv:2401.10208},
  year={2024}
}

@article{koh2023generating,
  title={Generating images with multimodal language models},
  author={Koh, Jing Yu and Fried, Daniel and Salakhutdinov, Russ R},
  journal={Advances in Neural Information Processing Systems},
  volume={36},
  pages={21487--21506},
  year={2023}
}

@article{bardes2024revisiting,
  title={Revisiting feature prediction for learning visual representations from video},
  author={Bardes, Adrien and Garrido, Quentin and Ponce, Jean and Chen, Xinlei and Rabbat, Michael and LeCun, Yann and Assran, Mahmoud and Ballas, Nicolas},
  journal={arXiv preprint arXiv:2404.08471},
  year={2024}
}

@online{worldlabs2025marble,
  author       = {{World Labs}},
  title        = {Marble: A Multimodal World Model},
  year         = {2025},
  month        = nov,
  day          = {12},
  url          = {https://www.worldlabs.ai/blog/marble-world-model},
  note         = {Accessed: 2026-02-22},
  organization = {World Labs}
}

@article{genie3,
  title         = {Genie 3: A New Frontier for World Models},
  author        = {Philip J. Ball and Jakob Bauer and Frank Belletti and Bethanie Brownfield and Ariel Ephrat and Shlomi Fruchter and Agrim Gupta and Kristian Holsheimer and Aleksander Holynski and Jiri Hron and Christos Kaplanis and Marjorie Limont and Matt McGill and Yanko Oliveira and Jack Parker-Holder and Frank Perbet and Guy Scully and Jeremy Shar and Stephen Spencer and Omer Tov and Ruben Villegas and Emma Wang and Jessica Yung and Cip Baetu and Jordi Berbel and David Bridson and Jake Bruce and Gavin Buttimore and Sarah Chakera and Bilva Chandra and Paul Collins and Alex Cullum and Bogdan Damoc and Vibha Dasagi and Maxime Gazeau and Charles Gbadamosi and Woohyun Han and Ed Hirst and Ashyana Kachra and Lucie Kerley and Kristian Kjems and Eva Knoepfel and Vika Koriakin and Jessica Lo and Cong Lu and Zeb Mehring and Alex Moufarek and Henna Nandwani and Valeria Oliveira and Fabio Pardo and Jane Park and Andrew Pierson and Ben Poole and Helen Ran and Tim Salimans and Manuel Sanchez and Igor Saprykin and Amy Shen and Sailesh Sidhwani and Duncan Smith and Joe Stanton and Hamish Tomlinson and Dimple Vijaykumar and Luyu Wang and Piers Wingfield and Nat Wong and Keyang Xu and Christopher Yew and Nick Young and Vadim Zubov and Douglas Eck and Dumitru Erhan and Koray Kavukcuoglu and Demis Hassabis and Zoubin Gharamani and Raia Hadsell and A{\"a}ron van den Oord and Inbar Mosseri and Adrian Bolton and Satinder Singh and Tim Rockt{\"a}schel},
  year          = {2025},
  url           = {}
}

@article{alonso2024diffusion,
  title={Diffusion for world modeling: Visual details matter in atari},
  author={Alonso, Eloi and Jelley, Adam and Micheli, Vincent and Kanervisto, Anssi and Storkey, Amos and Pearce, Tim and Fleuret, Fran{\c{c}}ois},
  journal={Advances in Neural Information Processing Systems},
  volume={37},
  pages={58757--58791},
  year={2024}
}

@inproceedings{
yang2024learning,
title={Learning Interactive Real-World Simulators},
author={Sherry Yang and Yilun Du and Seyed Kamyar Seyed Ghasemipour and Jonathan Tompson and Leslie Pack Kaelbling and Dale Schuurmans and Pieter Abbeel},
booktitle={The Twelfth International Conference on Learning Representations},
year={2024},
url={https://openreview.net/forum?id=sFyTZEqmUY}
}

@inbook{Ball_2020, place={Cambridge}, series={Cambridge Handbooks in Psychology}, title={Hypothetical Thinking}, booktitle={The Cambridge Handbook of the Imagination}, publisher={Cambridge University Press}, author={Ball, Linden J.}, editor={Abraham, AnnaEditor}, year={2020}, pages={514--528}, collection={Cambridge Handbooks in Psychology}}

@article{hafner2023mastering,
  title={Mastering diverse domains through world models},
  author={Hafner, Danijar and Pasukonis, Jurgis and Ba, Jimmy and Lillicrap, Timothy},
  journal={arXiv preprint arXiv:2301.04104},
  year={2023}
}

@article{ben2019demystifying,
  title={Demystifying parallel and distributed deep learning: An in-depth concurrency analysis},
  author={Ben-Nun, Tal and Hoefler, Torsten},
  journal={ACM Computing Surveys (CSUR)},
  volume={52},
  number={4},
  pages={1--43},
  year={2019},
  publisher={ACM New York, NY, USA}
}

@article{zhao2023pytorch,
  title={Pytorch fsdp: experiences on scaling fully sharded data parallel},
  author={Zhao, Yanli and Gu, Andrew and Varma, Rohan and Luo, Liang and Huang, Chien-Chin and Xu, Min and Wright, Less and Shojanazeri, Hamid and Ott, Myle and Shleifer, Sam and others},
  journal={arXiv preprint arXiv:2304.11277},
  year={2023}
}

@inproceedings{wu2025improved,
  title={Improved video vae for latent video diffusion model},
  author={Wu, Pingyu and Zhu, Kai and Liu, Yu and Zhao, Liming and Zhai, Wei and Cao, Yang and Zha, Zheng-Jun},
  booktitle={Proceedings of the Computer Vision and Pattern Recognition Conference},
  pages={18124--18133},
  year={2025}
}

@article{su2024roformer,
  title={Roformer: Enhanced transformer with rotary position embedding},
  author={Su, Jianlin and Ahmed, Murtadha and Lu, Yu and Pan, Shengfeng and Bo, Wen and Liu, Yunfeng},
  journal={Neurocomputing},
  volume={568},
  pages={127063},
  year={2024},
  publisher={Elsevier}
}

@article{duan2025worldscore,
  title={WorldScore: A Unified Evaluation Benchmark for World Generation},
  author={Duan, Haoyi and Yu, Hong-Xing and Chen, Sirui and Fei-Fei, Li and Wu, Jiajun},
  journal={arXiv preprint arXiv:2504.00983},
  year={2025}
}

@misc{worldmodelbench,
      title={WorldModelBench: Judging Video Generation Models As World Models}, 
      author={Dacheng Li and Yunhao Fang and Yukang Chen and Shuo Yang and Shiyi Cao and Justin Wong and Michael Luo and Xiaolong Wang and Hongxu Yin and Joseph E. Gonzalez and Ion Stoica and Song Han and Yao Lu},
      year={2025},
      eprint={2502.20694},
      archivePrefix={arXiv},
      primaryClass={cs.CV},
      url={https://arxiv.org/abs/2502.20694}, 
}

@misc{gao2025visionlanguagemodelsinternalworld,
  title={Do Vision-Language Models Have Internal World Models? Towards an Atomic Evaluation},
  author={Qiyue Gao and Xinyu Pi and Kevin Liu and Junrong Chen and Ruolan Yang and Xinqi Huang and Xinyu Fang and Lu Sun and Gautham Kishore and Bo Ai and Stone Tao and Mengyang Liu and Jiaxi Yang and Chao-Jung Lai and Chuanyang Jin and Jiannan Xiang and Benhao Huang and Zeming Chen and David Danks and Hao Su and Tianmin Shu and Ziqiao Ma and Lianhui Qin and Zhiting Hu},
  year={2025},
  eprint={2506.21876},
  archivePrefix={arXiv},
  primaryClass={cs.CL},
  url={https://arxiv.org/abs/2506.21876}
}

@misc{cosmos,
      title={Cosmos World Foundation Model Platform for Physical AI}, 
      author={NVIDIA},
      year={2025},
      eprint={2501.03575},
      archivePrefix={arXiv},
      primaryClass={cs.CV},
      url={https://arxiv.org/abs/2501.03575}, 
}

@online{nvidia-cosmos-predict2,
  title        = {Cosmos-Predict2: General-Purpose World Foundation Models for Physical AI},
  author       = {{NVIDIA Cosmos}},
  year         = {2025},
  url          = {https://github.com/nvidia-cosmos/cosmos-predict2},
  urldate      = {2025-10-06},
  note         = {GitHub repository}
}

@misc{OpenAI2025o3o4minisystemcard,
  author       = {OpenAI},
  title        = {OpenAI o3 and o4-mini System Card},
  year         = {2025},
  month        = {April},
  howpublished = {\url{https://cdn.openai.com/pdf/2221c875-02dc-4789-800b-e7758f3722c1/o3-and-o4-mini-system-card.pdf}},
  note         = {Accessed May 31, 2025}
}

@misc{openai2024gpt4technicalreport,
      title={GPT-4 Technical Report}, 
      author={OpenAI},
      year={2024},
      eprint={2303.08774},
      archivePrefix={arXiv},
      primaryClass={cs.CL},
      url={https://arxiv.org/abs/2303.08774}, 
}

@misc{assran2025vjepa2selfsupervisedvideo,
      title={V-JEPA 2: Self-Supervised Video Models Enable Understanding, Prediction and Planning}, 
      author={Mido Assran and Adrien Bardes and David Fan and Quentin Garrido and Russell Howes and Mojtaba and Komeili and Matthew Muckley and Ammar Rizvi and Claire Roberts and Koustuv Sinha and Artem Zholus and Sergio Arnaud and Abha Gejji and Ada Martin and Francois Robert Hogan and Daniel Dugas and Piotr Bojanowski and Vasil Khalidov and Patrick Labatut and Francisco Massa and Marc Szafraniec and Kapil Krishnakumar and Yong Li and Xiaodong Ma and Sarath Chandar and Franziska Meier and Yann LeCun and Michael Rabbat and Nicolas Ballas},
      year={2025},
      eprint={2506.09985},
      archivePrefix={arXiv},
      primaryClass={cs.AI},
      url={https://arxiv.org/abs/2506.09985}, 
}

@inproceedings{assran2023self,
  title={Self-supervised learning from images with a joint-embedding predictive architecture},
  author={Assran, Mahmoud and Duval, Quentin and Misra, Ishan and Bojanowski, Piotr and Vincent, Pascal and Rabbat, Michael and LeCun, Yann and Ballas, Nicolas},
  booktitle={Proceedings of the IEEE/CVF Conference on Computer Vision and Pattern Recognition},
  pages={15619--15629},
  year={2023}
}

@online{klingai-app-cn,
  title        = {Kling AI --- App Portal (Chinese)},
  author       = {{Kuaishou Technology}},
  year         = {2025},
  url          = {https://app.klingai.com/cn/},
  urldate      = {2025-10-06},
  note         = {Accessed 2025-10-06}
}

@online{hailuo-ai-minimax,
  title        = {Hailuo AI (MiniMax) --- Official Site},
  author       = {{MiniMax}},
  year         = {2025},
  url          = {https://hailuoaiminimax.com/},
  urldate      = {2025-10-06},
  note         = {Accessed 2025-10-06}
}

@online{runway-gen3-alpha,
  title   = {Introducing Gen-3 Alpha},
  author  = {{Runway Research}},
  year    = {2024},
  url     = {https://runwayml.com/research/introducing-gen-3-alpha},
  urldate = {2025-10-06},
  note    = {Accessed 2025-10-06}
}

@article{bu2025agibot,
  title={Agibot world colosseo: A large-scale manipulation platform for scalable and intelligent embodied systems},
  author={Bu, Qingwen and Cai, Jisong and Chen, Li and Cui, Xiuqi and Ding, Yan and Feng, Siyuan and Gao, Shenyuan and He, Xindong and Hu, Xuan and Huang, Xu and others},
  journal={arXiv preprint arXiv:2503.06669},
  year={2025}
}

@article{lynch2023interactive,
  title={Interactive language: Talking to robots in real time},
  author={Lynch, Corey and Wahid, Ayzaan and Tompson, Jonathan and Ding, Tianli and Betker, James and Baruch, Robert and Armstrong, Travis and Florence, Pete},
  journal={IEEE Robotics and Automation Letters},
  year={2023},
  publisher={IEEE}
}

@article{loshchilov2017decoupled,
  title={Decoupled weight decay regularization},
  author={Loshchilov, Ilya and Hutter, Frank},
  journal={arXiv preprint arXiv:1711.05101},
  year={2017}
}

@article{hao2023reasoning,
  title={ {Reasoning with Language Model is Planning with World Model} },
  author={Hao, Shibo and Gu, Yi and Ma, Haodi and Hong, Joshua Jiahua and Wang, Zhen and Wang, Daisy Zhe and Hu, Zhiting},
  journal={arXiv preprint arXiv:2305.14992},
  year={2023}
}

@article{xiang2023language,
  title={ {Language Models Meet World Models: Embodied Experiences Enhance Language Models} },
  author={Xiang, Jiannan and Tao, Tianhua and Gu, Yi and Shu, Tianmin and Wang, Zirui and Yang, Zichao and Hu, Zhiting},
  journal={arXiv preprint arXiv:2305.10626},
  year={2023}
}

@article{ha2018world,
  title={{World models}},
  author={Ha, David and Schmidhuber, J{\"u}rgen},
  journal={arXiv preprint arXiv:1803.10122},
  year={2018}
}

@article{battaglia2013simulation,
  title={{Simulation as an engine of physical scene understanding}},
  author={Battaglia, Peter W and Hamrick, Jessica B and Tenenbaum, Joshua B},
  journal={PNAS},
  year={2013}
}

@article{tolman1948cognitive,
  title={Cognitive maps in rats and men.},
  author={Tolman, Edward C},
  journal={Psychological review},
  volume={55},
  number={4},
  pages={189},
  year={1948},
  publisher={American Psychological Association}
}

@article{briscoe2011mental,
  title={Mental imagery and the varieties of amodal perception},
  author={Briscoe, Robert Eamon},
  journal={Pacific Philosophical Quarterly},
  volume={92},
  number={2},
  pages={153--173},
  year={2011},
  publisher={Wiley Online Library}
}

@article{schrittwieser2020mastering,
  title={Mastering atari, go, chess and shogi by planning with a learned model},
  author={Schrittwieser, Julian and Antonoglou, Ioannis and Hubert, Thomas and Simonyan, Karen and Sifre, Laurent and Schmitt, Simon and Guez, Arthur and Lockhart, Edward and Hassabis, Demis and Graepel, Thore and others},
  journal={Nature},
  volume={588},
  number={7839},
  pages={604--609},
  year={2020},
  publisher={Nature Publishing Group UK London}
}

@inproceedings{hafner2019learning,
  title={Learning latent dynamics for planning from pixels},
  author={Hafner, Danijar and Lillicrap, Timothy and Fischer, Ian and Villegas, Ruben and Ha, David and Lee, Honglak and Davidson, James},
  booktitle={International conference on machine learning},
  pages={2555--2565},
  year={2019},
  organization={PMLR}
}

@article{hafner2020mastering,
  title={Mastering atari with discrete world models},
  author={Hafner, Danijar and Lillicrap, Timothy and Norouzi, Mohammad and Ba, Jimmy},
  journal={arXiv preprint arXiv:2010.02193},
  year={2020}
}

@article{hafner2019dream,
  title={Dream to control: Learning behaviors by latent imagination},
  author={Hafner, Danijar and Lillicrap, Timothy and Ba, Jimmy and Norouzi, Mohammad},
  journal={arXiv preprint arXiv:1912.01603},
  year={2019}
}

@article{kingma2013auto,
  title={Auto-encoding variational {B}ayes},
  author={Kingma, Diederik P and Welling, Max},
  journal={arXiv preprint arXiv:1312.6114},
  year={2013}
}

@article{sutton2017reinforcement,
  title={Reinforcement learning: An introduction (2nd Edition)},
  author={Sutton, Richard S and Barto, Andrew G},
  year={2017},
  publisher={Cambridge, MA: MIT Press}
}

@inproceedings{vaswani2017attention,
  title={Attention is all you need},
  author={Vaswani, Ashish and Shazeer, Noam and Parmar, Niki and Uszkoreit, Jakob and Jones, Llion and Gomez, Aidan N and Kaiser, {\L}ukasz and Polosukhin, Illia},
  booktitle={NeurIPS},
  pages={5998--6008},
  year={2017}
}

@article{hu2023gaia,
  title={Gaia-1: A generative world model for autonomous driving},
  author={Hu, Anthony and Russell, Lloyd and Yeo, Hudson and Murez, Zak and Fedoseev, George and Kendall, Alex and Shotton, Jamie and Corrado, Gianluca},
  journal={arXiv preprint arXiv:2309.17080},
  year={2023}
}

@article{lecun2022path,
  title={A path towards autonomous machine intelligence version 0.9. 2, 2022-06-27},
  author={LeCun, Yann},
  journal={Open Review},
  volume={62},
  year={2022}
}

@misc{chen2024panda70m,
      title={Panda-70M: Captioning 70M Videos with Multiple Cross-Modality Teachers}, 
      author={Tsai-Shien Chen and Aliaksandr Siarohin and Willi Menapace and Ekaterina Deyneka and Hsiang-wei Chao and Byung Eun Jeon and Yuwei Fang and Hsin-Ying Lee and Jian Ren and Ming-Hsuan Yang and Sergey Tulyakov},
      year={2024},
      eprint={2402.19479},
      archivePrefix={arXiv},
      primaryClass={cs.CV}
}

@article{videoworldsimulators2024,
  title={Video generation models as world simulators},
  author={Tim Brooks and Bill Peebles and Connor Holmes and Will DePue and Yufei Guo and Li Jing and David Schnurr and Joe Taylor and Troy Luhman and Eric Luhman and Clarence Ng and Ricky Wang and Aditya Ramesh},
  year={2024},
  url={https://openai.com/research/video-generation-models-as-world-simulators},
}

@article{bruce2024genie,
  title={Genie: Generative Interactive Environments},
  author={Bruce, Jake and Dennis, Michael and Edwards, Ashley and Parker-Holder, Jack and Shi, Yuge and Hughes, Edward and Lai, Matthew and Mavalankar, Aditi and Steigerwald, Richie and Apps, Chris and others},
  journal={arXiv preprint arXiv:2402.15391},
  year={2024}
}

@article{wang2023drivedreamer,
  title={Drivedreamer: Towards real-world-driven world models for autonomous driving},
  author={Wang, Xiaofeng and Zhu, Zheng and Huang, Guan and Chen, Xinze and Lu, Jiwen},
  journal={arXiv preprint arXiv:2309.09777},
  year={2023}
}

@article{wang2023driving,
  title={Driving into the future: Multiview visual forecasting and planning with world model for autonomous driving},
  author={Wang, Yuqi and He, Jiawei and Fan, Lue and Li, Hongxin and Chen, Yuntao and Zhang, Zhaoxiang},
  journal={arXiv preprint arXiv:2311.17918},
  year={2023}
}

@article{zhao2024drivedreamer,
  title={DriveDreamer-2: LLM-Enhanced World Models for Diverse Driving Video Generation},
  author={Zhao, Guosheng and Wang, Xiaofeng and Zhu, Zheng and Chen, Xinze and Huang, Guan and Bao, Xiaoyi and Wang, Xingang},
  journal={arXiv preprint arXiv:2403.06845},
  year={2024}
}

@misc{zhou2024robodreamer,
      title={RoboDreamer: Learning Compositional World Models for Robot Imagination}, 
      author={Siyuan Zhou and Yilun Du and Jiaben Chen and Yandong Li and Dit-Yan Yeung and Chuang Gan},
      year={2024},
      eprint={2404.12377},
      archivePrefix={arXiv},
      primaryClass={cs.RO}
}

@misc{du2023learning,
      title={Learning Universal Policies via Text-Guided Video Generation}, 
      author={Yilun Du and Mengjiao Yang and Bo Dai and Hanjun Dai and Ofir Nachum and Joshua B. Tenenbaum and Dale Schuurmans and Pieter Abbeel},
      year={2023},
      eprint={2302.00111},
      archivePrefix={arXiv},
      primaryClass={cs.AI}
}

@misc{peebles2023scalable,
      title={Scalable Diffusion Models with Transformers}, 
      author={William Peebles and Saining Xie},
      year={2023},
      eprint={2212.09748},
      archivePrefix={arXiv},
      primaryClass={cs.CV}
}

@misc{sun2024generative,
      title={Generative Multimodal Models are In-Context Learners}, 
      author={Quan Sun and Yufeng Cui and Xiaosong Zhang and Fan Zhang and Qiying Yu and Zhengxiong Luo and Yueze Wang and Yongming Rao and Jingjing Liu and Tiejun Huang and Xinlong Wang},
      year={2024},
      eprint={2312.13286},
      archivePrefix={arXiv},
      primaryClass={cs.CV}
}

@misc{veo2024,
  author       = {Google DeepMind},
  title        = {Veo: A generative video model by Google DeepMind},
  howpublished = {\url{https://deepmind.google/models/veo/}},
  year         = {2025},
}

@article{yang2024cogvideox,
  title={Cogvideox: Text-to-video diffusion models with an expert transformer},
  author={Yang, Zhuoyi and Teng, Jiayan and Zheng, Wendi and Ding, Ming and Huang, Shiyu and Xu, Jiazheng and Yang, Yuanming and Hong, Wenyi and Zhang, Xiaohan and Feng, Guanyu and others},
  journal={arXiv preprint arXiv:2408.06072},
  year={2024}
}

@article{kong2024hunyuanvideo,
  title={Hunyuanvideo: A systematic framework for large video generative models},
  author={Kong, Weijie and Tian, Qi and Zhang, Zijian and Min, Rox and Dai, Zuozhuo and Zhou, Jin and Xiong, Jiangfeng and Li, Xin and Wu, Bo and Zhang, Jianwei and others},
  journal={arXiv preprint arXiv:2412.03603},
  year={2024}
}

@article{wan2025wan,
  title={Wan: Open and advanced large-scale video generative models},
  author={Wan, Team and Wang, Ang and Ai, Baole and Wen, Bin and Mao, Chaojie and Xie, Chen-Wei and Chen, Di and Yu, Feiwu and Zhao, Haiming and Yang, Jianxiao and others},
  journal={arXiv preprint arXiv:2503.20314},
  year={2025}
}

@article{parker2024genie,
  title={Genie 2: A large-scale foundation world model},
  author={Parker-Holder, J and Ball, P and Bruce, J and Dasagi, V and Holsheimer, K and Kaplanis, C and Moufarek, A and Scully, G and Shar, J and Shi, J and others},
  journal={URL: https://deepmind. google/discover/blog/genie-2-a-large-scale-foundation-world-model},
  year={2024}
}

@misc{worldlabs2024generating,
  author       = {{World Labs}},
  title        = {Generating Worlds -- World Labs},
  year         = {2024},
  month        = dec,
  url          = {https://www.worldlabs.ai/blog/generating-worlds}
}

@article{feng2024matrix,
  title={The matrix: Infinite-horizon world generation with real-time moving control},
  author={Feng, Ruili and Zhang, Han and Yang, Zhantao and Xiao, Jie and Shu, Zhilei and Liu, Zhiheng and Zheng, Andy and Huang, Yukun and Liu, Yu and Zhang, Hongyang},
  journal={arXiv preprint arXiv:2412.03568},
  year={2024}
}

@article{bai2025qwen2,
  title={Qwen2. 5-vl technical report},
  author={Bai, Shuai and Chen, Keqin and Liu, Xuejing and Wang, Jialin and Ge, Wenbin and Song, Sibo and Dang, Kai and Wang, Peng and Wang, Shijie and Tang, Jun and others},
  journal={arXiv preprint arXiv:2502.13923},
  year={2025}
}

@article{chung2023unimax,
  title={Unimax: Fairer and more effective language sampling for large-scale multilingual pretraining},
  author={Chung, Hyung Won and Constant, Noah and Garcia, Xavier and Roberts, Adam and Tay, Yi and Narang, Sharan and Firat, Orhan},
  journal={arXiv preprint arXiv:2304.09151},
  year={2023}
}

@article{betker2023improving,
  title={Improving image generation with better captions},
  author={Betker, James and Goh, Gabriel and Jing, Li and Brooks, Tim and Wang, Jianfeng and Li, Linjie and Ouyang, Long and Zhuang, Juntang and Lee, Joyce and Guo, Yufei and others},
  journal={Computer Science. https://cdn. openai. com/papers/dall-e-3. pdf},
  volume={2},
  number={3},
  pages={8},
  year={2023}
}

@misc{zeng2023mixnetaccuratedetectionchallenging,
      title={MixNet: Toward Accurate Detection of Challenging Scene Text in the Wild}, 
      author={Yu-Xiang Zeng and Jun-Wei Hsieh and Xin Li and Ming-Ching Chang},
      year={2023},
      eprint={2308.12817},
      archivePrefix={arXiv},
      primaryClass={cs.CV},
      url={https://arxiv.org/abs/2308.12817}, 
}

@inproceedings{murray2012ava,
  title={AVA: A large-scale database for aesthetic visual analysis},
  author={Murray, Naila and Marchesotti, Luca and Perronnin, Florent},
  booktitle={CVPR},
  year={2012},
}

@misc{schuhmann2022improved_aesthetic_predictor,
  author       = {Schuhmann, Christoph},
  title        = {Improved Aesthetic Predictor: CLIP+MLP Aesthetic Score Predictor},
  year         = {2022},
  howpublished = {\url{https://github.com/christophschuhmann/improved-aesthetic-predictor}},
  note         = {Apache 2.0 licensed; starred 1.1k times on GitHub (accessed 2025-06-26)}
}

@inproceedings{esser2024scaling,
  title={Scaling rectified flow transformers for high-resolution image synthesis},
  author={Esser, Patrick and Kulal, Sumith and Blattmann, Andreas and Entezari, Rahim and M{\"u}ller, Jonas and Saini, Harry and Levi, Yam and Lorenz, Dominik and Sauer, Axel and Boesel, Frederic and others},
  booktitle={Forty-first international conference on machine learning},
  year={2024}
}

@article{shah2024flashattention,
  title={Flashattention-3: Fast and accurate attention with asynchrony and low-precision},
  author={Shah, Jay and Bikshandi, Ganesh and Zhang, Ying and Thakkar, Vijay and Ramani, Pradeep and Dao, Tri},
  journal={Advances in Neural Information Processing Systems},
  volume={37},
  pages={68658--68685},
  year={2024}
}

@article{dong2024flex,
  title={Flex attention: A programming model for generating optimized attention kernels},
  author={Dong, Juechu and Feng, Boyuan and Guessous, Driss and Liang, Yanbo and He, Horace},
  journal={arXiv preprint arXiv:2412.05496},
  year={2024}
}

@article{li2021sequence,
  title={Sequence parallelism: Long sequence training from system perspective},
  author={Li, Shenggui and Xue, Fuzhao and Baranwal, Chaitanya and Li, Yongbin and You, Yang},
  journal={arXiv preprint arXiv:2105.13120},
  year={2021}
}

@article{jacobs2023deepspeed,
  title={Deepspeed ulysses: System optimizations for enabling training of extreme long sequence transformer models},
  author={Jacobs, Sam Ade and Tanaka, Masahiro and Zhang, Chengming and Zhang, Minjia and Song, Shuaiwen Leon and Rajbhandari, Samyam and He, Yuxiong},
  journal={arXiv preprint arXiv:2309.14509},
  year={2023}
}

@inproceedings{zhang2023adding,
  title={Adding conditional control to text-to-image diffusion models},
  author={Zhang, Lvmin and Rao, Anyi and Agrawala, Maneesh},
  booktitle={Proceedings of the IEEE/CVF international conference on computer vision},
  pages={3836--3847},
  year={2023}
}

@article{zhang2025sageattention2++,
  title={Sageattention2++: A more efficient implementation of sageattention2},
  author={Zhang, Jintao and Xu, Xiaoming and Wei, Jia and Huang, Haofeng and Zhang, Pengle and Xiang, Chendong and Zhu, Jun and Chen, Jianfei},
  journal={arXiv preprint arXiv:2505.21136},
  year={2025}
}

@article{ho2022classifier,
  title={Classifier-free diffusion guidance},
  author={Ho, Jonathan and Salimans, Tim},
  journal={arXiv preprint arXiv:2207.12598},
  year={2022}
}

@article{lipman2022flow,
  title={Flow matching for generative modeling},
  author={Lipman, Yaron and Chen, Ricky TQ and Ben-Hamu, Heli and Nickel, Maximilian and Le, Matt},
  journal={arXiv preprint arXiv:2210.02747},
  year={2022}
}

@article{xing2025critiques,
  title={Critique of World Model},
  author={Xing, Eric and Deng, Mingkai and Hou, Jinyu},
  journal={arXiv preprint arXiv:2507.05169},
  year={2025}
}

@article{decart2024oasis,
  title={Oasis: A universe in a transformer},
  author={Decart, Etched and McIntyre, Quinn and Campbell, Spruce and Chen, Xinlei and Wachen, Robert},
  journal={URL: https://oasis-model. github. io},
  year={2024}
}

@article{guo2025mineworld,
  title={Mineworld: a real-time and open-source interactive world model on minecraft},
  author={Guo, Junliang and Ye, Yang and He, Tianyu and Wu, Haoyu and Jiang, Yushu and Pearce, Tim and Bian, Jiang},
  journal={arXiv preprint arXiv:2504.08388},
  year={2025}
}

@article{liu2022flow,
  title={Flow straight and fast: Learning to generate and transfer data with rectified flow},
  author={Liu, Xingchao and Gong, Chengyue and Liu, Qiang},
  journal={arXiv preprint arXiv:2209.03003},
  year={2022}
}

@inproceedings{zhoudenoising,
  title={Denoising Diffusion Bridge Models},
  author={Zhou, Linqi and Lou, Aaron and Khanna, Samar and Ermon, Stefano},
  booktitle={The Twelfth International Conference on Learning Representations}
}

@inproceedings{zhoudino,
  title={DINO-WM: World Models on Pre-trained Visual Features enable Zero-shot Planning},
  author={Zhou, Gaoyue and Pan, Hengkai and LeCun, Yann and Pinto, Lerrel},
  booktitle={Forty-second International Conference on Machine Learning}
}

@article{abu2016youtube,
  title={Youtube-8m: A large-scale video classification benchmark},
  author={Abu-El-Haija, Sami and Kothari, Nisarg and Lee, Joonseok and Natsev, Paul and Toderici, George and Varadarajan, Balakrishnan and Vijayanarasimhan, Sudheendra},
  journal={arXiv preprint arXiv:1609.08675},
  year={2016}
}

@article{zellers2021merlot,
  title={Merlot: Multimodal neural script knowledge models},
  author={Zellers, Rowan and Lu, Ximing and Hessel, Jack and Yu, Youngjae and Park, Jae Sung and Cao, Jize and Farhadi, Ali and Choi, Yejin},
  journal={Advances in neural information processing systems},
  volume={34},
  pages={23634--23651},
  year={2021}
}

@misc{nvidia_cosmos_bench,
      title={Cosmos-Reason1: From Physical Common Sense To Embodied Reasoning}, 
      author={NVIDIA and : and Alisson Azzolini and Junjie Bai and Hannah Brandon and Jiaxin Cao and Prithvijit Chattopadhyay and Huayu Chen and Jinju Chu and Yin Cui and Jenna Diamond and Yifan Ding and Liang Feng and Francesco Ferroni and Rama Govindaraju and Jinwei Gu and Siddharth Gururani and Imad El Hanafi and Zekun Hao and Jacob Huffman and Jingyi Jin and Brendan Johnson and Rizwan Khan and George Kurian and Elena Lantz and Nayeon Lee and Zhaoshuo Li and Xuan Li and Maosheng Liao and Tsung-Yi Lin and Yen-Chen Lin and Ming-Yu Liu and Xiangyu Lu and Alice Luo and Andrew Mathau and Yun Ni and Lindsey Pavao and Wei Ping and David W. Romero and Misha Smelyanskiy and Shuran Song and Lyne Tchapmi and Andrew Z. Wang and Boxin Wang and Haoxiang Wang and Fangyin Wei and Jiashu Xu and Yao Xu and Dinghao Yang and Xiaodong Yang and Zhuolin Yang and Jingxu Zhang and Xiaohui Zeng and Zhe Zhang},
      year={2025},
      eprint={2503.15558},
      archivePrefix={arXiv},
      primaryClass={cs.AI},
      url={https://arxiv.org/abs/2503.15558}, 
}

\end{document}